\documentclass[letterpaper]{article}

\usepackage{arxiv}
\usepackage{standalone}

\usepackage{amsmath}
\usepackage{amsfonts}
\usepackage{amssymb}

\usepackage{algorithm}
\usepackage{algpseudocode}

\usepackage[utf8]{inputenc} %
\usepackage[T1]{fontenc}    %
\usepackage[hidelinks]{hyperref}
\usepackage[nameinlink,capitalize]{cleveref}
\usepackage{url}            %
\usepackage{booktabs}       %

\usepackage{nicefrac}       %
\usepackage{microtype}      %
\usepackage{graphicx}
\usepackage{caption}
\usepackage{multirow}
\usepackage[numbers]{natbib}
\usepackage{doi}

\usepackage[section]{placeins} %
\usepackage{float}              %

\usepackage[dvipsnames]{xcolor}
\usepackage{makecell}
\usepackage[version=4]{mhchem}
\usepackage[per-mode=symbol,round-precision=3,scientific-notation=false,range-phrase = \text{--}]{siunitx}
\usepackage{wrapfig}
\usepackage{adjustbox}

\usepackage{xspace}%

\newcommand{\drref}[1]{\hyperref[dr:#1]{DR#1}}

\makeatletter
\newcommand*{\etc}{%
    \@ifnextchar{.}%
        {etc}%
        {etc.\@\xspace}%
}
\makeatother

\newcommand{\myCOMMENT}[1]{}

\DeclareUnicodeCharacter{025B}{\ensuremath{\varepsilon}}

\graphicspath{{figures/}}

\usepackage{listings}

\definecolor{bkgd}{RGB}{240,242,246}
\definecolor{ceruleanblue}{rgb}{0.16, 0.32, 0.75}
\definecolor{orange-red}{rgb}{1.0, 0.27, 0.0}
\definecolor{anotherblue}{RGB}{37,92,243}
\definecolor{blackblue}{RGB}{46,60,85}
\definecolor{goldyellow}{RGB}{199,146,12}
\lstdefinestyle{altstyle2}{
    backgroundcolor=\color{bkgd},
    basicstyle=\ttfamily\footnotesize\color{blackblue},
    breakatwhitespace=false,
    breaklines=true,
    captionpos=b,
    commentstyle=\color{goldyellow},
    keepspaces=true,
    keywordstyle=\color{orange-red},
    language=Python,
    numbersep=5pt,
    numberstyle=\tiny\color{ceruleanblue},
    showspaces=false,
    showstringspaces=false,
    showtabs=false,
    stringstyle=\color{anotherblue},
    tabsize=2,
    numbers=left
}
\lstdefinelanguage{json}{
    basicstyle=\ttfamily\footnotesize\color{blackblue},
    string=[s]{"}{"},
    stringstyle=\color{anotherblue},
    comment=[l]{//},
    morecomment=[s]{/*}{*/},
    commentstyle=\color{goldyellow},
    literate=
        *{:}{{{\color{orange-red}:}}}{1}
        {,}{{{\color{orange-red},}}}{1}
        {\{}{{{\color{ceruleanblue}\{}}}{1}
        {\}}{{{\color{ceruleanblue}\}}}}{1}
        {[}{{{\color{ceruleanblue}[}}}{1}
        {]}{{{\color{ceruleanblue}]}}}{1}
        {true}{{{\color{orange-red}true}}}{4}
        {false}{{{\color{orange-red}false}}}{5}
        {null}{{{\color{orange-red}null}}}{4},
}

\usepackage[scaled=0.86]{helvet}

\crefname{lstlisting}{listing}{listings}
\Crefname{lstlisting}{Listing}{Listings}

\crefname{algorithm}{Alg.}{Algs.}
\Crefname{algorithm}{Alg.}{Algs.}

\usepackage{pdflscape}

\algnewcommand{\Inputs}[1]{%
  \State \textbf{Inputs:}
  \Statex \hspace*{\algorithmicindent}\parbox[t]{.8\linewidth}{\raggedright #1}
}
\algnewcommand{\Initialize}[1]{%
  \State \textbf{Initialize:}
  \Statex \hspace*{\algorithmicindent}\parbox[t]{.8\linewidth}{\raggedright #1}
}

\definecolor{darkorange}{HTML}{FF8C00}
\newcommand{\cf}[1]{%
  \ifnum#1>100%
    \textcolor{darkorange}{#1}%
  \else%
    \ifnum#1>75%
      \textcolor{black}{#1}%
    \else%
      \textcolor{teal}{#1}%
    \fi%
  \fi%
}

\definecolor{textred}{RGB}{170,0,0}

\usepackage{eqparbox}

\usepackage{tikz}
\usetikzlibrary{shapes.geometric, arrows, arrows.meta, positioning, calc, matrix, fit, backgrounds}
\usepackage{pgfplots}
\pgfplotsset{compat=1.18}
\usepgfplotslibrary{groupplots}
\usetikzlibrary{patterns}

\usepackage{tcolorbox}
\tcbuselibrary{skins}

\definecolor{regmsgfr}{HTML}{0072B2}       %
\definecolor{appmsgfr}{HTML}{009E73}       %
\definecolor{hitlchkfr}{HTML}{E69F00}      %
\definecolor{ledgerclr}{HTML}{8B959E}      %

\definecolor{badgepub}{HTML}{009E73}       %
\definecolor{badgeprop}{HTML}{D55E00}      %
\definecolor{badgesgi}{HTML}{750014}       %
\definecolor{badgecui}{HTML}{FF1423}       %

\newcommand{\classbadge}[2]{%
  \tikz[baseline=(b.base)]{%
    \node[fill=#1, rounded corners=2pt, inner xsep=3pt, inner ysep=1.5pt,
          font=\sffamily\scriptsize\bfseries\color{white}] (b) {#2};}%
}
\newcommand{\cpub}{\classbadge{badgepub}{PUBLIC}}
\newcommand{\cprop}{\classbadge{badgeprop}{PROPRIETARY}}
\newcommand{\csgi}{\classbadge{badgesgi}{SGI}}
\newcommand{\ccui}{\classbadge{badgecui}{CUI}}

\tcbset{
  reg msg/.style={
    enhanced, colback=regmsgfr!8, colframe=regmsgfr,
    colbacktitle=regmsgfr, coltitle=white, fonttitle=\footnotesize\bfseries,
    left=4pt, right=4pt, top=4pt, bottom=3pt,
    arc=2pt, outer arc=2pt, boxrule=0.6pt,
    fontupper=\footnotesize,
    before skip=0pt, after skip=0pt,
  },
  app msg/.style={
    enhanced, colback=appmsgfr!8, colframe=appmsgfr,
    colbacktitle=appmsgfr, coltitle=white, fonttitle=\footnotesize\bfseries,
    left=4pt, right=4pt, top=4pt, bottom=3pt,
    arc=2pt, outer arc=2pt, boxrule=0.6pt,
    fontupper=\footnotesize,
    before skip=0pt, after skip=0pt,
  },
}

\newcommand{\regmsg}[3]{%
  \par\vspace{6pt}\noindent
  \begin{minipage}{0.78\linewidth}%
    \begin{tcolorbox}[reg msg,
      title={\textbf{Regulator Agent}\hfill\texttt{#1}\quad#2}]
      #3
    \end{tcolorbox}%
  \end{minipage}\par
}

\newcommand{\appmsg}[3]{%
  \par\vspace{6pt}\noindent\hspace*{0.22\linewidth}%
  \begin{minipage}{0.78\linewidth}%
    \begin{tcolorbox}[app msg,
      title={\textbf{Applicant Agent}\hfill\texttt{#1}\quad#2}]
      #3
    \end{tcolorbox}%
  \end{minipage}\par
}

\newcommand{\ledgerlog}[1]{%
  \par\vspace{2pt}%
  {\centering\textcolor{ledgerclr}{\small\dotfill\;\textit{#1}\;\dotfill}\par}%
  \vspace{2pt}%
}

\newcommand{\hitlcheck}[1]{%
  \par\vspace{4pt}%
  \noindent\hspace*{0.225\linewidth}%
  \begin{minipage}{0.55\linewidth}%
    \begin{tcolorbox}[
      enhanced, colback=hitlchkfr!10, colframe=hitlchkfr,
      left=4pt, right=4pt, top=3pt, bottom=3pt,
      arc=2pt, boxrule=0.8pt,
      borderline={0.4pt}{1.5pt}{hitlchkfr, dashed},
      fontupper=\small\sffamily, halign=center,
      before skip=0pt, after skip=0pt,
    ]
      \textbf{Human On The Loop Checkpoint}\\#1
    \end{tcolorbox}%
  \end{minipage}\par
  \vspace{4pt}%
}

\title{Overcoming the Regulatory Bottleneck via Agent-to-Agent Protocols: A Nuclear Case Study}

\author{ 
	Akshay J.~Dave \\
	Nuclear Science and Engineering\\
	Argonne National Laboratory\\
	Lemont, IL 60439\\
	\texttt{ajd@anl.gov} \\
	\And
	David Grabaskas \\
	Nuclear Science and Engineering\\
	Argonne National Laboratory\\
	Lemont, IL 60439\\
	\And
	Joseph A.~Renevitz \\
	Idaho National Laboratory\\
  1955 N.~Fremont Ave.\\
  Idaho Falls, ID 83415
	\And
	Richard B.~Vilim \\
	Nuclear Science and Engineering\\
	Argonne National Laboratory\\
	Lemont, IL 60439\\
}

\date{}

\renewcommand{\headeright}{ }
\renewcommand{\undertitle}{ }
\renewcommand{\shorttitle}{Overcoming the Regulatory Bottleneck via Agent-to-Agent Protocols}

\begin{document}
\maketitle

\begin{abstract}

Regulatory review of advanced nuclear reactor designs routinely spans more than three years and consumes hundreds of millions of dollars in combined regulator and applicant labor. We present the Regulatory Context Protocol (RCP), an Agent-to-Agent communication standard that replaces the formal human-to-human pipeline between regulators and applicants with a structured, auditable agentic channel, while preserving human oversight at safety-significant decision points. The protocol is calibrated against an analysis of 1{,}236 documents from U.S. Nuclear Regulatory Commission advanced reactor dockets and demonstrated with a working multi-agent pilot. Against an \$89M, 42-month Reconstructed Baseline, RCP cuts costs by 50--77\% (\$21M--\$44M) and timelines by 65\% (15~months). Without a shared protocol, Standalone Agents reach only \$54M--\$74M and 21~months. The residual cost-and-time gap is structural, not algorithmic: it traces to the inter-organizational pipeline that only an agent-to-agent standard can compress. The same bottleneck---formal multi-party review under strict auditability requirements---characterizes pharmaceutical approvals, environmental permitting, financial supervision, and aviation certification. The US regulatory paperwork burden carries a \$426.5~billion annual opportunity cost; replicated broadly, the projected 50--77\% reduction implies savings on the order of \$210--\$330~billion per year---approaching 1\% of US GDP.

\vspace{1.0em}

\end{abstract}

\keywords{\centering Regulatory Compliance Automation \and Multi-Agent Systems \and Large Language Models \and AI Governance \and Agent-to-Agent Protocols \and E-Government}

\section{Introduction}

The contemporary administrative state faces a structural misalignment between the exponential complexity of the regulated economy and the linear capacity of manual verification workflows. This paper establishes that the crisis of scale is a fundamental limit of applying 20th-century processes to a 21st-century data environment, necessitating new technological approaches.

The regulatory burden in the United States has reached significant levels. As of September 2024, the government-wide paperwork burden was estimated at 12.1~billion hours~\cite{AAF2024burden}. Over the preceding 14-month period, federal requirements added approximately 1.6~billion hours to this total. Economically, using Bureau of Labor Statistics wage data, this time represents an opportunity cost of approximately \$426.5~billion~\cite{AAF2024burden}.

While the Paperwork Reduction Act (PRA) and initiatives like the Biden Administration's Burden Reduction Initiative aim to manage this growth, the volume of new data continues to outpace manual reduction efforts. For example, FY~2023 saw reduction claims of 34.5~million hours against a backdrop of larger programmatic increases~\cite{AAF2024burden}. This disparity suggests that linear interventions (simplifying forms) may face limits in solving what is effectively an exponential data scaling problem.

The complexity of compliance is further compounded by the volume of regulatory text. The Mercatus Center's QuantGov project tracks regulatory restrictions---instances of mandatory language such as shall, must, or required. At the federal level, the Code of Federal Regulations (CFR) exceeds 100~million words containing approximately 1.1~million restrictions~\cite{Mercatus2024}. The manual verification model assumes a human agent can comprehensively recall and apply this vast library, a task that increasingly exceeds effective human information processing capacity.

Traditional manual workflows, such as those used in nuclear licensing, are thus becoming bottlenecks for innovation and deployment. In this paper, we propose the Regulatory Context Protocol (RCP), a cross-industry framework leveraging multi-agent Large Language Models (LLMs) to automate and strictly structure regulatory interactions. To demonstrate its efficacy, we present a case study in the nuclear sector, automating the Request for Additional Information (RAI) process using the U.S. Nuclear Regulatory Commission's (NRC) Agencywide Documents Access and Management System (ADAMS) as a foundation.

\subsection{The Throughput Bottleneck}

The aggregate regulatory volume translates into operational throughput challenges, particularly in agencies managing high-stakes technical reviews. We examine specific sectors to demonstrate that this is a systemic scalability failure driven by reliance on manual verification.

\textit{Patent and Trademark Office (USPTO).} The backlog of unexamined patent applications reached 793,824 in FY~2024. However, the crisis is most visible in the Request for Continued Examination (RCE) loop. When a manual examiner rejects a claim due to ambiguity, the applicant files an RCE, placing the file back into the queue. In FY~2024, the inventory of pending RCEs contributed to an average total pendency of 30~months for such applications~\cite{USPTO2024trends}. This churn represents a failure of the initial verification interface; the system relies on correction by rejection rather than interactive validation, stretching total pendency for complex applications significantly~\cite{USPTO2024trends}.

\textit{Food and Drug Administration (FDA).} The Office of Generic Drugs saw full approvals drop by 11.2\% in FY~2024, with First-Cycle Approval rates stuck below 18\%~\cite{FDA2024generics}. Beyond documentation, the manual inspection model has collapsed post-COVID. Nearly 2,000 drug manufacturing plants (42\% of the registered universe) were overdue for safety checks as of 2024~\cite{APfda2024}. The agency's pivot to a risk-based approach acknowledges a hard truth: it lacks the human capacity to fulfill its statutory mandate of comprehensive physical verification.

\textit{The Legacy of Paper (IRS and OPM).} The scalability crisis is exacerbated by the persistence of analog workflows. In the 2024 filing season, the Internal Revenue Service (IRS) spent \$8.65 to process a paper return compared to just \$0.23 for an electronic return---a 37-fold cost differential~\cite{GAOIRS2024}. Similarly, the Office of Personnel Management (OPM) persistently misses retirement processing goals due to its ``continuing reliance on paper-based applications and manual processing''~\cite{GAOOPM2019}. These failures are not managerial but structural; they represent the Human Capital Ceiling, where the complexity of compliance exceeds the available supply of trained human verifiers.

\subsection{The Nuclear Licensing Bottleneck}

The nuclear sector represents a distinct case of these dynamics, where the consequences of the crisis of scale are most acute. While the NRC has maintained rigorous oversight of the operating fleet, the transition to licensing advanced non-light water reactors (non-LWRs) has highlighted the interface challenges between legacy frameworks and new technologies~\cite{NRCReqRoadmap}. This structural misalignment is particularly evident in the application of 10~CFR~Part~50~\cite{CFRPart50} and Part~52~\cite{CFRPart52} frameworks, which were originally optimized for large, established light-water designs. When applied to novel, smaller designs, these comprehensive standards create significant adaptation friction. As noted in the NRC's Regulatory Review Roadmap, designers face considerable difficulty navigating the options for pre-application review versus formal licensing~\cite{NRCReqRoadmap}. This structural mismatch forces a heavy reliance on manual interpretation to bridge the gap between deterministic regulations and novel safety cases. The NRC's recently finalized 10~CFR Part~53---a risk-informed, technology-inclusive framework developed specifically for advanced reactors---directly acknowledges this mismatch and modernizes the criteria against which designs are evaluated~\cite{NRCPart53}. However, Part~53 reforms \textit{what} is reviewed; it does not change \textit{how} the review is conducted. The underlying interaction model remains a document-mediated, manually-verified exchange between applicant and regulator, leaving the throughput bottleneck this paper targets fundamentally intact.

These friction points result in an extremely high cost of information exchange. NuScale Power, for instance, invested over \$500~million and 2~million labor hours to complete its Design Certification Application (DCA)~\cite{NuScaleCost}. A key driver of this cost is the phenomenon of design finalization during the review process. The NRC's recent lessons learned report highlighted that the design evolved significantly during the review—such as the redesign of the emergency core cooling system valves—which triggered massive rework and re-review cycles~\cite{NuScaleLL2022}. This iteration loop, where manual verification must effectively restart after every design change, acts as a primary bottleneck inhibiting rapid deployment.

Beyond structural and process mismatch, friction also arises from ambiguous regulatory language and operational silos. The NuScale review highlighted the lack of a clear definition for credible events; without a precise, computable definition, applicants and regulators spend valuable cycles debating whether an event requires mitigation~\cite{NuScaleLL2022}. This unpredictability increases review durations and costs, as interpretations can vary among staff—a classic symptom of a system relying on distributed human consensus rather than formalized verification. Furthermore, the NRC's traditional matrixed review structure, where technical branches review specific chapters in isolation, struggles to evaluate holistic safety cases. While the agency has recommended interdisciplinary review teams to better handle novel, integrated features~\cite{NuScaleLL2022}, implementing such structural changes remains a slow process.

These challenges are exacerbated by severe human capital limits. The NRC competes with industry for talent, with an average time-to-hire of 148~days in FY~2023~\cite{OIGHiring2024}. This human capital ceiling implies that adding more manual verifiers is not a viable scaling strategy; the agency cannot hire fast enough to match the complexity of the incoming advanced reactor wave. Economic factors further compound this pressure. Unlike most agencies, the NRC is mandated to recover approximately 100\% of its budget from fees, resulting in a professional hourly rate of \textasciitilde\$318~\cite{FeeRule2025}. While the ADVANCE Act of 2024~\cite{ADVANCEAct2024} lowers rates for advanced reactors to mitigate this, the underlying economics suggest that moving from manual verification to automated, scalable assurance is essential for both the regulator and the applicant.

\subsection{AI Agents in the Public Sector}\label{sec:ai_public_sector}

Federal agencies have already begun adopting AI to attack the bottlenecks identified above, though deployments to date remain bounded within individual organizations. The USPTO released its enterprise \textit{Artificial Intelligence Strategy} in January~2025~\cite{USPTOAIStrategy} and, in October~2025, launched an \textit{Automated Search Pilot Program} that uses an AI-powered tool to surface up to ten ranked prior-art references for an application before it reaches an examiner~\cite{USPTOAutoSearch}. The NRC's own \textit{Artificial Intelligence Strategic Plan} (NUREG-2261) commits the agency to AI deployment across knowledge management, operating-experience analysis, and regulatory-decision support through FY~2027, and explicitly anticipates license applications incorporating AI technologies in the near term~\cite{NRCAIStrategy}.

This activity sits within an updated federal policy stance. Executive Order~14179, ``Removing Barriers to American Leadership in Artificial Intelligence'' (January~2025), revokes the prior administration's EO~14110 and directs agencies to accelerate AI use in mission-critical workflows~\cite{EO14179}. In parallel, Executive Order~14300 (May~2025) orders the reform of the NRC, specifically mandating a maximum 18-month timeline for new reactor licensing reviews~\cite{EO14300}. Office of Management and Budget (OMB) Memorandum~M-25-21 (April~2025) operationalizes this directive: it rescinds and replaces M-24-10 while preserving its core compliance perimeter, requiring mandatory human oversight, AI impact assessments, and continuous monitoring for ``High-Impact AI''---systems whose outputs serve as a principal basis for decisions with significant legal, material, or safety effect~\cite{OMBM2521}. Any production-grade regulatory AI must satisfy these requirements.

Yet every deployment cited above stops at the organizational boundary. Each agency builds AI for use on \textit{its own} documents, by \textit{its own} staff, against \textit{its own} internal data---mirroring the silos of legacy regulatory IT systems. The interface \textit{between} organizations---where applicants and regulators must exchange formally docketed information across trust boundaries---remains a manual, document-mediated workflow. As the throughput and verification bottlenecks above make clear, this is precisely where review cost and schedule are concentrated: NuScale's \$500~million and 4{,}000~RAIs were not generated by inefficient internal review, but by the friction of formal inter-organizational exchange under strict auditability constraints~\cite{NuScaleCost,NuScaleLL2022}. Internal automation alone cannot resolve a bottleneck whose locus is the interaction \textit{between} two organizations. What is missing is a standardized, protocol-driven channel through which agency and applicant agents can communicate directly while preserving the legal accountability, privacy, and auditability that statutory regulatory processes require---the gap that RCP, presented in the next section, is designed to fill.

\subsection{Objectives and Outline}

The objective of this paper is to establish a rigorous, cross-industry framework for automating regulatory interactions using multi-agent LLMs, while preserving the legal accountability, privacy, and transparency requirements inherent to safety-critical domains. Specifically, we aim to formalize the regulatory interface as a secure, shared-record construction problem with strict privacy constraints, define RCP as a domain profile of existing agent communication standards, and demonstrate the framework's viability through a case study in nuclear licensing. To this end, \cref{sec:framework} presents the complete RCP methodology, beginning with the foundational communication protocols for internal knowledge access and inter-agent messaging, followed by the four Design Requirements derived from the ``crisis of scale,'' the system architecture, trust boundaries, and protocol specification that realize them. \Cref{sec:case_study} applies the framework to nuclear regulatory automation, describing the mining of regulatory patterns from the NRC ADAMS database, the implementation of hierarchical agent architectures for both regulator and applicant, and a walkthrough demonstration of an automated RAI cycle. \Cref{sec:results_policy} evaluates the framework's performance in terms of efficiency gains, enhanced transparency through the Context Stream, and mechanisms for handling sensitive data classifications. Finally, \cref{sec:conclusion} summarizes contributions and discusses future work toward cross-agency adoption.

\begin{figure}[!ht]
    \centering
    \hspace*{-1.25pt}\includegraphics{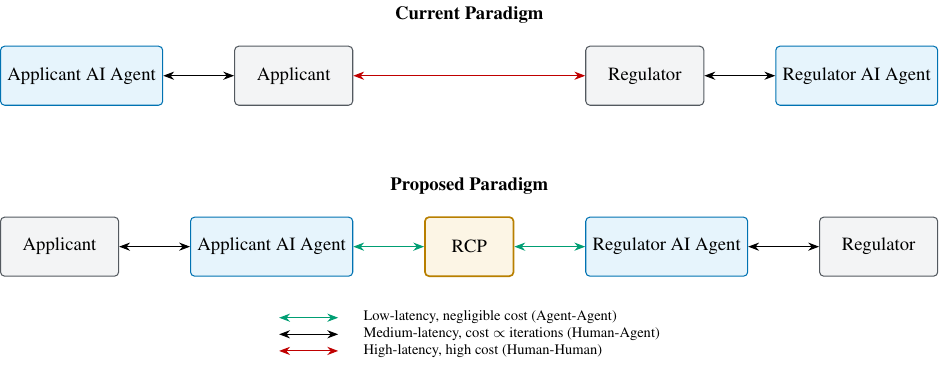}
    \caption{RCP automation flow, contrasting the current human-bottlenecked paradigm (top) with the proposed agent-mediated paradigm under RCP (bottom).}
    \label{fig:rcp_flow}
\end{figure}

\Cref{fig:rcp_flow} states the central hypothesis of this paper visually: the current paradigm (top) routes every inter-organizational exchange through a human-to-human pipeline whose throughput is bounded by reviewer staff hours; the proposed paradigm (bottom) replaces that pipeline with protocol-driven agent-to-agent communication under the RCP state protocol, while reserving human intervention for high-level intent and final safety determinations. The remainder of the paper develops the protocol that realizes this transformation and quantifies its cost and schedule consequences.

\section{The Regulatory Context Protocol Framework}\label{sec:framework}

Reframing the regulator--applicant interface is the central move of this paper. The 4{,}000~RAIs and \$500~million cited above are not the cost of writing documents; they are the cost of two organizations trying to keep a shared technical record consistent across a trust boundary, under strict legal auditability, while each protects information the other is not entitled to see. Because the applicant and regulator act as independent entities with private memory coordinating over a network, this is fundamentally a \textit{distributed protocol} problem. Treating it as such---rather than as a document-processing pipeline---admits engineering primitives (typed messages, state machines, append-only ledgers) designed natively for cross-boundary coordination. This section develops the framework: foundational communication protocols, formal design methodology and requirements, and the system architecture that realize them.

\subsection{Foundational Communication Protocols}

Multi-agent systems for regulatory automation require two distinct communication layers: one for connecting agents to internal knowledge sources, and another for inter-agent collaboration across organizational boundaries. Specialized agents collaborate within strictly defined boundaries---for example, a legal research agent retrieves a statute while a compliance agent verifies a claim~\cite{GaaSframework}---and ``reason through problems and take autonomous action'' only within those bounds~\cite{AgenticState}. This decomposition is what makes agents tractable in a safety-critical setting: each agent completes a discrete step, ``which allows errors to be traced back''~\cite{AspireMultiAgent}, producing an automated audit trail that directly supports the High-Impact AI oversight requirements established by OMB~M-25-21 (\cref{sec:ai_public_sector}). The remainder of this section introduces the two protocols that realize this two-layer architecture and explains how RCP profiles them for regulatory use.

\subsubsection{Model Context Protocol}
The Model Context Protocol (MCP) is an open standard defining a uniform interface for AI agents to invoke external Tools, read external Resources, and use shared Prompts through a small set of typed primitives~\cite{MCP2024}. By replacing the combinatorial tangle of bespoke model-to-system integrations with a single protocol, MCP exposes each accessible tool and data source through a schema-defined contract that the agent discovers at runtime---grounding its reasoning in declared, validated resources rather than ad-hoc connector code.

\subsubsection{Agent-to-Agent Protocol}
The Agent-to-Agent (A2A) protocol is an open standard for direct communication between independent agents, whether they run side-by-side on a single host, within an organization, or across the public internet~\cite{A2A2024}. The protocol defines four composable primitives: \textit{Agent Cards}, \textit{Messages}, \textit{Tasks}, and \textit{Artifacts}. Because the entire framework of this paper rests on A2A, we introduce each at the abstract level here; the RCP profile that specializes them with regulatory semantics is given in \cref{sec:protocol_spec}.

\textit{Agent Cards} are the identity and capability advertisements that make discovery possible. A Card is a machine-readable descriptor declaring an agent's identifier, the categories of work it accepts, the message types it understands, and any authentication scopes that bound its authority. Before any work begins, two agents exchange Cards to determine whether they have the authority and capability to interact at all, in the same way that parties to a formal proceeding establish their credentials and standing before substantive exchange begins.

\textit{Messages} are the typed payloads exchanged between agents. Each Message declares a type drawn from an agreed vocabulary, carries a structured body conforming to that type's schema, and includes routing and provenance metadata. Messages are not free-form chat. Because the type and structure of a Message are declared in machine-readable fields rather than embedded in prose, any receiving agent---or any intermediary along the path, such as a router, audit logger, or access-control filter---can determine what action the Message requires by reading those fields directly. Routing, validation, and downstream dispatch become deterministic operations on declared structure rather than probabilistic inferences over natural language, eliminating a class of failures in which the intent of an unstructured request is misclassified or its required follow-up is missed.

\textit{Tasks} are stateful units of work that organize Messages into a coherent interaction. A Task has a defined lifecycle---submitted, working, completed, with branches for failure, pending input, or required human authorization---tracked independently by each participating agent. A Task may be short-lived (a single request-response) or long-running, spanning many messages and intermediate state transitions before reaching a terminal state.

\textit{Artifacts} are the immutable outputs that a Task produces. An Artifact is a discrete file or structured data object that constitutes the durable record of completed work. Artifacts are content-addressable: a cryptographic hash binds the object to its identity, making post-hoc tampering detectable and producing a verifiable chain of custody from the input messages to the final output.

The primitives compose hierarchically: Agent Cards define \textit{who} can interact, Tasks define \textit{what} is being done, Messages constitute the \textit{exchange} within a Task, and Artifacts constitute its \textit{output}. Where MCP standardizes how an agent reaches \textit{into} a system, A2A standardizes how agents reach \textit{each other}.

\subsubsection{RCP as a Domain Profile}
RCP is not a replacement for these standards, but a \textit{domain profile} that specializes A2A for regulatory interactions. RCP constrains A2A's generic primitives with regulatory semantics: Agent Cards declare docket authorization, Messages carry legally-significant payloads, Tasks enforce human sovereignty checkpoints, and Artifacts become immutable evidence. This layered approach enables regulatory agents to leverage standard A2A tooling while enforcing the strict requirements of safety-critical licensing workflows. Together, MCP and A2A form the ``intranet'' and ``internet'' of the RCP framework, shifting the verification model from unstructured chat to logged, protocol-driven exchange.

\subsection{Design Requirements}\label{sec:design_requirements}

The regulator--applicant interface is, at base, a problem of co-constructing a verifiable shared record across a trust boundary; we now make that framing precise enough to derive engineering requirements from it. We model each party as holding a private knowledge base---$S_A$ for the applicant (design data, proprietary methods, internal compliance reasoning) and $S_R$ for the regulator (deliberations, precedents, pre-decisional analysis)---together with a shared, append-only docket $C$ that constitutes the legally binding record of the proceeding. Progress consists of bounded, signed disclosures from one private state into $C$; everything else stays behind the trust boundary. From this picture, and from the cost and capacity constraints established in Section~1, we derive four Design Requirements (DRs) that any candidate protocol must satisfy.

\phantomsection\label{dr:1}\textbf{DR1: Information Sovereignty.}
Regulatory interactions involve trade secrets, export-controlled information, Safeguards Information, and pre-decisional deliberations whose unauthorized disclosure carries legal and competitive consequences.
\textit{Constraint:} Each organization must retain unilateral control over which subsets of $S_A$ or $S_R$ enter $C$, and over the cryptographic provenance of every disclosure it authors. The protocol must make it structurally impossible for one party to read or compute over the other's private state without an explicit, signed disclosure into the shared record. Compute infrastructure may be cloud-hosted or on-premises; the requirement is on access control and cryptographic attestation of disclosure, not on deployment topology.

\phantomsection\label{dr:2}\textbf{DR2: Verifiable, Discoverable Shared State.}
Probabilistic language models cannot serve as the system of record for safety-significant decisions, and administrative law independently requires a reviewable record sufficient to support adjudication, dispute resolution, and legal discovery---including under the ex parte rules of 10~CFR~2.347 and~2.348, before the Atomic Safety and Licensing Board (ASLB), and under the Freedom of Information Act (FOIA).
\textit{Constraint:} The Context Stream $C$ must be a deterministic, signed, append-only record in which every entry is cryptographically signed by a named principal and is content-addressable by hash, so that the record cannot be silently revised and so that $C$---rather than any agent's transient memory---is the source of truth for downstream review. The protocol must additionally ensure that \textit{every} message crossing the trust boundary between applicant and regulator agents enters $C$, including informal Pattern~A exchanges, with sufficient metadata (sender identity, recipient identity, timestamp, classification, parent task) to support both real-time oversight and post-hoc legal discovery. No backchannel may exist outside the protocol's logging path; classification labels (\cref{sec:protocol_spec}) determine \textit{who} may read each entry, but every entry must exist in some auditable form.

\phantomsection\label{dr:3}\textbf{DR3: Epistemic Grounding.}
Cryptographic immutability protects the integrity of the record, but not its correctness; a signed hallucination is a permanent hallucination. Because LLMs can produce fluent but unsupported claims, the protocol cannot rely on agent self-reports for substantive technical assertions.
\textit{Constraint:} Every binding agent-authored entry to $C$ must be traceable to a citation chain in declared knowledge sources accessible through MCP---regulatory text, controlled engineering documents, or signed prior artifacts in $C$ itself. Entries that cannot be grounded must be flagged for human authorship rather than auto-generated, and the grounding chain must itself be appended to the record so that the basis of every regulator-facing claim is auditable in its own right.

\phantomsection\label{dr:4}\textbf{DR4: Human Oversight.}
Regulatory authority is delegated by statute to accountable human officials, not to algorithms; agents may propose and draft, but only humans may commit binding actions to the record.
\textit{Constraint:} Human oversight is continuous rather than episodic. The full Context Stream is observable by authorized human reviewers in real time, and any reviewer may intervene at any point in any task. Within this continuous regime, a designated subset of state transitions---those producing a legally binding act, such as the issuance of a formal RAI, the submission of a docketed application, or the release of a Safety Evaluation Report---require an explicit cryptographic signature from a named human principal before they may commit. The protocol must distinguish drafted entries from signed entries at the message level, and must refuse to advance binding tasks past the human checkpoint without that signature.

\subsection{RCP Specification}\label{sec:protocol_spec}

RCP standardizes what crosses the organizational trust boundary. Every message exchanged between an applicant and a regulator must be an RCP message, signed by the sending agent and written to a shared, signed, append-only Context Stream $C$ hosted on neutral infrastructure called the RCP Host. The Host routes messages and persists them in $C$; it never receives any organization's private state. Inside an organization, an agent reaches into its own private knowledge---proprietary databases, analysis codes, pre-decisional documents---through whatever interface the organization chooses; in practice MCP is the canonical option, since it provides a standard, schema-validated layer over internal tools and resources, but RCP is intentionally agnostic about internal access. This division is what enforces Information Sovereignty (\drref{1}): the only protocol that crosses the boundary is RCP, and an RCP message carries only what the sending organization's agent explicitly authored. The result is a system in which the \textit{process} is fully transparent and evidential (\drref{2}) while each organization's underlying private state is never directly exposed to the other party. The protocol does not eliminate the possibility that an agent authors a message that discloses more than it should; what it does guarantee is that every cross-boundary message carries a sensitivity classification (\cref{sec:privacy_security}) and is routed under handling rules that match its label, so that even inadvertent disclosures are tagged, logged, and access-controlled rather than silent.

\begin{figure}[!ht]
    \centering
    \hspace*{-1.25pt}\includegraphics{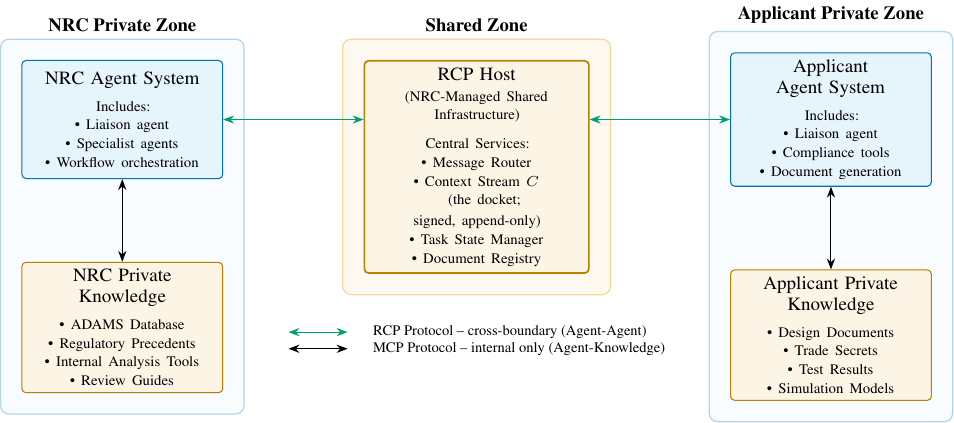}
    \caption{RCP component architecture, separating each organization's private zone (with its own knowledge bases and agents) from the shared RCP Host that routes cross-boundary messages and persists them in the Context Stream $C$.}
    \label{fig:component_arch}
\end{figure}

\Cref{fig:component_arch} expands the high-level paradigm shift of \cref{fig:rcp_flow} into a component-level view of this division. Each organization's private zone houses its own knowledge bases, agents, and internal tools (typically accessed via MCP), all of which remain inside the boundary. The shared RCP Host carries the Context Stream $C$ and the routing layer that enforces classification, signature, and human-checkpoint constraints on every RCP message crossing between zones.

The remainder of this section specifies the precise mechanics of RCP as a domain profile of the A2A standard~\cite{A2A2024}. Rather than defining a new protocol, RCP constrains and semantically extends A2A's generic primitives---Agent Cards, Messages, Contexts, Tasks, Artifacts, and Security Schemes---for the regulatory domain. We organize the specification by conceptual layer: identity, communication, workflow, and data governance.

\subsubsection{Agent Identity and Discovery}\label{sec:agent_identity}
The first step in any RCP interaction is mutual, authenticated discovery. A2A provides a standard mechanism for this: every agent publishes a machine-readable identity record---the \texttt{AgentCard}---at a well-known location, describing who the agent is, what kinds of work it accepts, and how it expects callers to authenticate themselves~\cite{A2A2024}. A counterpart retrieves this record before any substantive communication and uses it to decide whether the peer has the standing and capabilities the interaction requires. RCP overrides the AgentCard at two points: the protocol profile a participating agent must declare, and the authorization vocabulary under which it accepts calls.

\textit{Protocol profile declaration.} A2A allows an agent to advertise the named protocol profiles it implements. RCP requires that every conformant agent advertise its support for a specific, versioned RCP profile, so that two agents can confirm at handshake time that they share the same regulatory semantics; agents whose declared profiles are incompatible refuse to begin substantive communication. This requirement also gives the standard a controlled evolution path: as RCP advances, an agent's declared version tells the network exactly which schemas and behaviors it implements, without out-of-band coordination. Within the profile declaration, an RCP agent further specifies its organizational role (applicant or regulator), the docket numbers it is authorized to act on, and the endpoints to which human-oversight escalations should be routed (\drref{4}). This structured identity matters most in nuclear licensing, where the NRC employs a matrixed review structure: separate domain specialists in thermal-hydraulics, structural engineering, neutronics, and other disciplines review specific chapters of a Safety Analysis Report (SAR) in parallel. The Agent Card makes this organizational expertise legible to the protocol, enabling automated routing and identification of cross-cutting issues---a direct response to the coordination challenges documented in advanced reactor licensing reviews~\cite{NuScaleLL2022}.

\textit{Authorization vocabulary.} A2A specifies how agents \textit{authenticate} (proving who they are) but leaves \textit{authorization} (what an authenticated caller may do) to each deployment, via a generic security-scheme slot in the AgentCard. RCP fills this slot with a fixed vocabulary of three named permissions: \textit{read} (passive docket monitoring), \textit{write} (sending messages or artifacts within an existing task), and \textit{workflow initiation} (opening a new regulatory task such as an RAI cycle). Every call carries one or more, and the receiving agent rejects any call whose permissions do not cover its operation. In our demonstration, the applicant and regulator agents each hold \textit{write} and \textit{workflow initiation} on the demonstration docket. Token \textit{issuance}---who decides that a given agent legitimately holds a given permission---remains the responsibility of each organization's identity infrastructure; what RCP fixes is the vocabulary itself, so that authorization checks are uniform network-wide. This is what allows \drref{1} to be enforced by the protocol rather than by per-agent code.

\subsubsection{Regulatory Communication Semantics}
Once agents have discovered each other, they communicate using A2A \texttt{Message} objects---the atomic unit of inter-agent communication, transporting structured content in an array of typed parts (text, file, or data). RCP profiles this primitive by enforcing a closed vocabulary of message types that mirror legally distinct communicative acts in the licensing process. Each message type carries structured metadata---including the regulatory basis (specific CFR citations such as 10~CFR~50.46~\cite{CFR50_46}) and the technical domain---validated against a formal schema. This distinction matters because NRC interactions carry different legal weight: a lightweight \texttt{information-request} for technical clarification is categorically different from a formal \texttt{rai-notification} that initiates the regulatory response clock. \Cref{tab:rcp_messages} enumerates the primary message types defined in the specification.

\begin{table}[htbp]
    \centering
    \caption{RCP Message Type Catalog defining the domain-specific communicative acts.}
    \label{tab:rcp_messages}
    \begin{tabular}{@{}llcl@{}}
        \toprule
        \textbf{Workflow} & \textbf{Message Type} & \textbf{Direction} & \textbf{Description} \\ \midrule
        \multirow{2}{*}{Info Exchange} & \texttt{information-request} & R $\to$ A & Query for technical data \\
        & \texttt{information-response} & A $\to$ R & Structured data response \\ \midrule
        \multirow{2}{*}{Document} & \texttt{document-submission} & A $\to$ R & Formal artifacts \\
        & \texttt{document-acknowledgment} & R $\to$ A & Cryptographic receipt \\ \midrule
        \multirow{3}{*}{RAI Cycle} & \texttt{rai-notification} & R $\to$ A & Formal issuance \\
        & \texttt{rai-content} & R $\to$ A & Regulatory questions \\
        & \texttt{rai-response} & A $\to$ R & Verified response \\ \bottomrule
    \end{tabular}
\end{table}

A2A also defines a \texttt{Context} primitive: a session identifier carried by each message and used to group related interactions into a coherent thread. RCP requires that this identifier be the official NRC docket number, so that the protocol's grouping mechanism aligns directly with the docket as it is already recognized in regulatory law. The consequence of this binding is that the Context Stream $C$ \textit{is} the docket: the same record viewed through the protocol lens (as the append-only message log under the Context identifier) and through the regulatory lens (as the official record of the proceeding). We use ``Context Stream'' when discussing the protocol-level data structure and ``docket'' when invoking its legal status; they refer to the same artifact. By construction, every message exchanged between an applicant and a regulator agent is bound to a docket; there is no protocol path by which two agents can exchange substantive information outside it. This forecloses the backchannel and gives effect to \drref{2}: the ``truth'' of a proceeding is defined by the docket, not by any agent's transient memory.

The docket itself is persistent, accumulating every cross-boundary message and artifact for the full duration of a review---typically months to years. It is a system of record, not an agent's prompt context: agents retrieve from the docket on demand, loading only the parent task, cited artifacts, and immediately relevant prior messages into per-turn working memory, so the docket can grow without bound while individual agent calls remain scoped to the task at hand. Persistence is itself a legal requirement: the docket is subject to adjudicatory hearings before the ASLB, public scrutiny through FOIA requests, and ex parte controls under 10~CFR~2.347 and~2.348, and must remain self-contained long after the original reviewers have moved on. Our corpus analysis underscores why completeness matters---over 90\% of the regulatory interactions analyzed were part of multi-step sequences rather than standalone exchanges, so partial records would be evidentially incoherent.

\subsubsection{Task Lifecycle and Human Oversight}
Messages alone do not constitute a regulatory proceeding---they must be organized into auditable workflows with deterministic execution. A2A defines a \texttt{Task} object as a state machine governing task progression from submission through completion or failure, with intermediate states for awaiting input and authorization~\cite{A2A2024}. RCP fixes the \textit{lifecycle semantics} that any such task must obey but leaves the workflow vocabulary---the specific kinds of regulatory tasks each agent supports---to the deploying organization, since the appropriate granularity depends on the licensing pathway and the agency's review structure. The substantive constraint RCP imposes at this layer is to reinterpret A2A's generic \texttt{auth-required} state as the cryptographically enforced human-signature checkpoint required by \drref{4} for binding regulatory actions.

\begin{figure}[!ht]
    \centering
    \hspace*{-1.25pt}\includegraphics{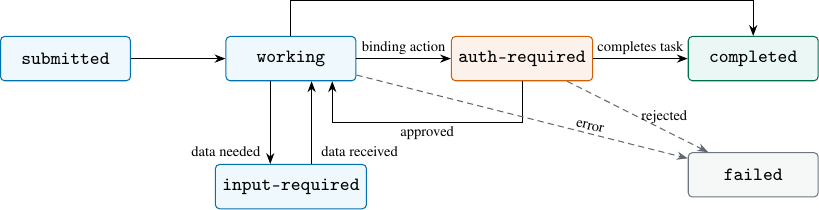}
    \caption{RCP task state machine. Routine work transitions from \texttt{submitted} through \texttt{working} to \texttt{completed}, optionally pausing in \texttt{input-required} while awaiting a counterparty response. Binding regulatory actions must pass the \texttt{auth-required} gate, where a cryptographic human signature (\drref{4}) is required before the task may complete; rejected or errored tasks terminate in \texttt{failed}.}
    \label{fig:task_fsm}
\end{figure}

\Cref{fig:task_fsm} illustrates the RCP task state machine. For routine verification tasks---formatting checks, cross-referencing, or information retrieval---the state machine proceeds from \texttt{submitted} through \texttt{working} to \texttt{completed}, transiting the \texttt{input-required} state whenever an agent must pause for a clarification or data response from its counterpart and resuming once that input arrives. None of this path requires human authorization, so routine tasks proceed asynchronously and in parallel. Any transition that would result in a binding regulatory commitment, however, must pass through the \texttt{auth-required} state. This checkpoint models specific points of human accountability in the NRC workflow: an applicant agent preparing to submit proprietary design data must pause until the responsible officer provides a signature corresponding to the withholding justification required under 10~CFR~2.390~\cite{CFR2_390}; an NRC agent's draft RAI must await approval before formal issuance initiates the regulatory response deadline. At this checkpoint, the agent generates a human-escalation message containing a summary of the proposed action. The task cannot advance to \texttt{completed} until a cryptographic signature from an authorized human is received, ensuring that the legal act of submission remains an auditable human responsibility. If the human rejects the proposed action, the task transitions to \texttt{failed} with a rejection rationale.

Tasks produce \texttt{Artifact} objects---discrete outputs such as regulatory documents, analysis reports, or formal submissions. RCP profiles artifacts by mandating content-addressable storage using SHA-256 hashes and defining a taxonomy of artifact types specific to licensing (e.g., regulatory documents, RAI packages). This ensures that every regulatory document is immutable, verifiable, and semantically classified---satisfying the Verifiable Shared State requirement (\drref{2}) by establishing a cryptographic chain of custody for all artifacts produced during the proceeding.

\subsubsection{Data Classification and Access Control}\label{sec:data_classification}
Beyond workflow control, the protocol must govern the sensitivity of exchanged data. RCP embeds data classification directly into the message structure by requiring agents to pre-mark all data with one of four classification labels before transmission:
\begin{itemize}
    \item \cpub{} --- Information not subject to withholding under any of the categories below; the default classification when no other applies, encompassing publicly accessible regulatory text, published technical literature, and applicant material the originator has not marked for withholding.
    \item \cprop{} --- Trade secrets and commercial or financial information protected from disclosure under 10~CFR~2.390~\cite{CFR2_390}.
    \item \csgi{} --- Safeguards Information subject to enhanced physical and access controls.
    \item \ccui{} --- Controlled Unclassified Information with security-related handling requirements.
\end{itemize}

This design directly addresses a significant operational burden in current practice: the manual, error-prone redaction of proprietary or security-sensitive content from documents destined for the public ADAMS docket. Classification is enforced at the source, within the Private Zone (via MCP), at the time data is first ingested---not retroactively applied after the fact. The protocol schema enforces this by requiring a withholding justification for any data marked \cprop{} (per 10~CFR~2.390(a)(4)) and recommending the generation of a public summary---a sanitized version suitable for public release. Classification is thus an intrinsic, validated property of the data object rather than an external metadata tag. This source-level enforcement, combined with the trust boundary separation illustrated in \cref{fig:component_arch}, ensures that the disclosure path from each agent's private state $S$ into the shared record $C$ respects sensitivity constraints: agents cannot inadvertently transmit data at a higher classification than the channel permits. Classification thus complements the authorization vocabulary introduced in \cref{sec:agent_identity}: where authorization scopes gate \textit{which operations} an agent may perform on the docket, classification labels gate \textit{which audiences} each message may reach.

\section{Case Study: Nuclear Licensing Process Automation}\label{sec:case_study}

To ground the abstract primitives of the RCP framework---such as cryptographic trust boundaries and append-only Context Streams---in real-world regulatory dynamics, we applied it to the nuclear licensing domain. Advanced reactor licensing represents an ideal stress test for distributed protocol automation: proceedings involve immense data volumes, stringent safety-critical oversight, and a statutory requirement to shield proprietary commercial designs (under 10~CFR~2.390) from the public record. In this section, we transition from theory to practice by first extracting empirical interaction patterns from the NRC's ADAMS database, then mapping those patterns into hierarchical agent architectures, and finally demonstrating an end-to-end automated RAI cycle using a representative advanced reactor subsystem.

\subsection{Regulatory Corpus Analysis}\label{sec:corpus_analysis}

To inform the design of RCP message types and specialist agent architectures, we conducted an empirical analysis of the NRC's ADAMS corpus. This analysis aimed to identify the most common interaction patterns, technical domains, and regulatory bases cited in real licensing proceedings.

\subsubsection{Dataset and Methodology}
Our initial corpus focused on advanced reactor applications---specifically documents associated with NuScale, Holtec, X-Energy, Kairos, and other advanced reactor applicants, spanning both Part~52 design certification and Part~50 construction permit pathways. This subset was chosen to capture the regulatory patterns most relevant to the next generation of licensing reviews. Documents were retrieved from ADAMS in September~2025 and constitute the analysis cutoff used throughout this paper; the full ADAMS database contains significantly more data spanning the entire operating fleet under 10~CFR~Part~50, which represents a richer corpus for future expansion. The analysis pipeline processed 1,236 regulatory documents including RAI responses, audit reports, topical reports, and formal submissions. All inference was performed locally using OpenAI's open-weight \texttt{gpt-oss-120b} model~\cite{OpenAIgptoss2025} served via \texttt{ollama}~\cite{Ollama}.

To structure the extracted information, we employed a hybrid taxonomy combining a fixed schema with adaptive expansion. Static fields impose strict validation on categories whose value sets are bounded and well-established---for example, \texttt{risk\_significance} (high/medium/low) and \texttt{regulatory\_pathway} (10~CFR Part 50/52/70/72)---ensuring consistent aggregation across the corpus. Adaptive fields, by contrast, permit the model to introduce new category values as it encounters them, with periodic consolidation to retire near-duplicates; these include \texttt{technical\_domains}, \texttt{action\_type}, and \texttt{blocking\_issues}. The adaptive component captures domain-specific distinctions that would be impractical to enumerate \emph{a priori}, such as the separation of ``Awaiting NRC Guidance'' from ``Awaiting Vendor Data'' as distinct blocking conditions, without requiring manual curation of the taxonomy.

\subsubsection{Key Findings}

\begin{figure}[!ht]
    \centering
    \hspace*{-1.25pt}\includegraphics{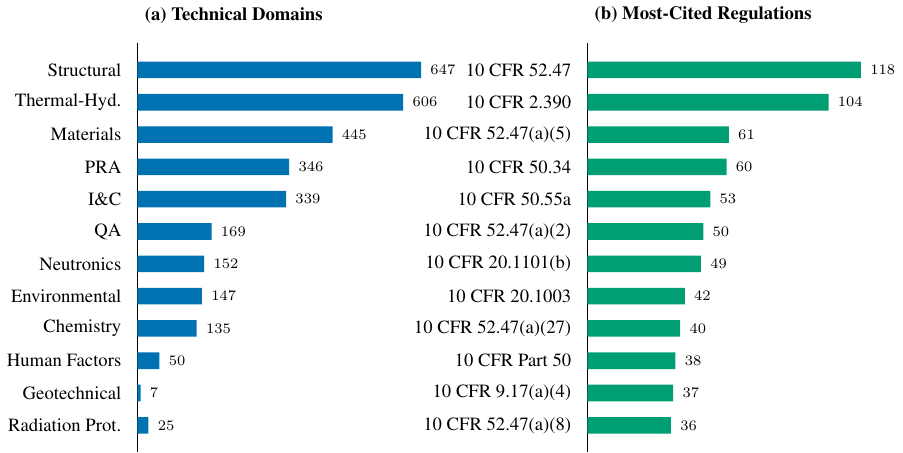}
    \caption{Distribution of technical domains (panel~(a)) and most-cited regulations (panel~(b)) across 1,236 advanced reactor regulatory documents. The dominance of Structural and Thermal-Hydraulics domains informed the design of specialist agents; the prevalence of 10~CFR~52 citations reflects the Part~52 pathway focus.}
    \label{fig:adams_domains_regs}
\end{figure}

Two views of the corpus characterize where regulatory effort concentrates (\cref{fig:adams_domains_regs}). The technical-domain distribution (panel~(a)) is dominated by Structural engineering (647 documents) and Thermal-Hydraulics (606), reflecting the centrality of these areas to safety analysis; Probabilistic Risk Assessment appears prominently as well (346), consistent with the NRC's risk-informed posture. On the regulatory-basis side (panel~(b)), 10~CFR~52.47~\cite{CFR52_47} (Standard Design Certification contents) leads with 118 citations, reflecting the Part~52 design certification pathway that dominates advanced reactor applications.

\begin{figure}[!ht]
    \centering
    \includegraphics{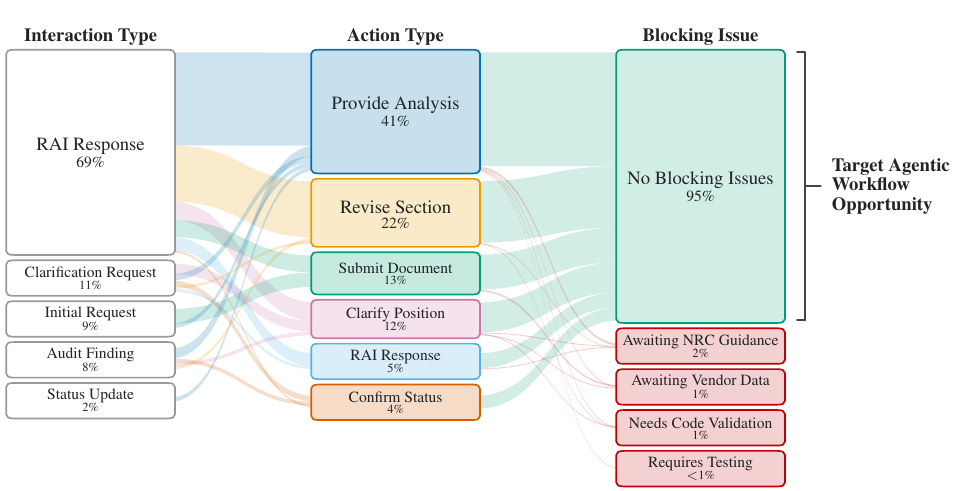}
    \caption{Regulatory document workflow from interaction type through action type to blocking issue. Flow widths are proportional to document counts; colors indicate destination.}
    \label{fig:adams_sankey}
\end{figure}

The end-to-end flow of regulatory interactions reveals where automation is most tractable. Tracing the relationships between interaction types, action types, and downstream blocking issues, RAI Responses account for the majority of documents (69\%), most often resolving into ``Provide Analysis'' or ``Revise Section'' actions. Critically, 95\% of documents proceed without blocking issues---the target opportunity for agentic workflow automation. The remaining 5\% encounter blockers such as ``Awaiting NRC Guidance'' or ``Awaiting Vendor Data,'' external dependencies that require human coordination and would not be eliminated by automating the in-band workflow. \Cref{fig:adams_sankey} renders these flows as a Sankey diagram.

\subsubsection{Implications for RCP Design}
The corpus analysis' distributions and flow structure provide concrete priors that shape four design decisions in the RCP framework and any agentic system built on top of it. First, the \textit{message catalog} (\cref{tab:rcp_messages}) is sized to the observed interaction inventory. Distinguishing \texttt{rai-content}, \texttt{information-request}, and \texttt{document-submission} as first-class message types---rather than collapsing them into a single ``review correspondence'' channel---is justified by the empirical separation between formal RAIs, lightweight clarifications, and submission events in the corpus. This typing is what allows downstream systems to apply different policies---for example, restricting human-on-the-loop (HOTL) gating to \texttt{rai-content} alone to satisfy \drref{4}: Human Oversight---without parsing free-text intent.

Second, the technical-domain distribution gives a basis for \textit{specialist agent decomposition and orchestration}. The long-tailed distribution---a small number of high-volume domains (Structural, Thermal-Hydraulics, Materials, PRA) carrying most of the traffic, with many low-volume domains in the tail---argues against either a single generalist reviewer or one specialist per CFR subsection. A practical orchestration pattern is to route head-of-distribution domains to dedicated specialists with curated tools and evaluation suites, while delegating tail domains to a shared generalist with broader retrieval scope. Liaison-side routing can use the \texttt{technical\_domains} and \texttt{regulatory\_basis} fields directly, avoiding an LLM-based classifier in the hot path for traffic that is already structurally typed.

Third, the regulatory-basis distribution informs \textit{retrieval-augmented generation (RAG) corpus design} for the agents themselves. The dominance of 10~CFR Part~52 citations and a small set of recurring CFR sections means that RAG indices for advanced-reactor review do not need to span the full Title~10 corpus to be effective; a tiered index with the top-cited sections (and their associated regulatory guides and SRP chapters) embedded at higher fidelity, and the long tail kept at coarser granularity, will cover the bulk of queries. Equally important, the \texttt{regulatory\_basis} field in incoming RCP messages can be used as a hard pre-filter on the retrieval set, narrowing the candidate corpus before semantic search and reducing both latency and the surface for hallucinated citations---directly supporting the Epistemic Grounding requirement (\drref{3}).

Fourth, the workflow analysis bounds where automation can credibly contribute. The 95\%/5\% split between non-blocked and externally blocked documents is the empirical justification for Pattern~A (lightweight exchange) as a structurally separate path from Pattern~B (formal RAI): the bulk of traffic is amenable to in-band agentic resolution, while the residual blockers (``Awaiting NRC Guidance,'' ``Awaiting Vendor Data'') are coordination problems that no protocol can dissolve. By routing these lightweight exchanges through the protocol, RCP ensures they are immutably logged, satisfying the Verifiable Shared State requirement (\drref{2}). This split also frames a meaningful evaluation metric for deployed RCP systems---the fraction of interactions resolved through Pattern~A without escalation---rather than the less informative aggregate of ``messages exchanged.''

\subsection{Implementation}
\begin{figure}[!ht]
    \centering
    \hspace*{-1.25pt}\includegraphics{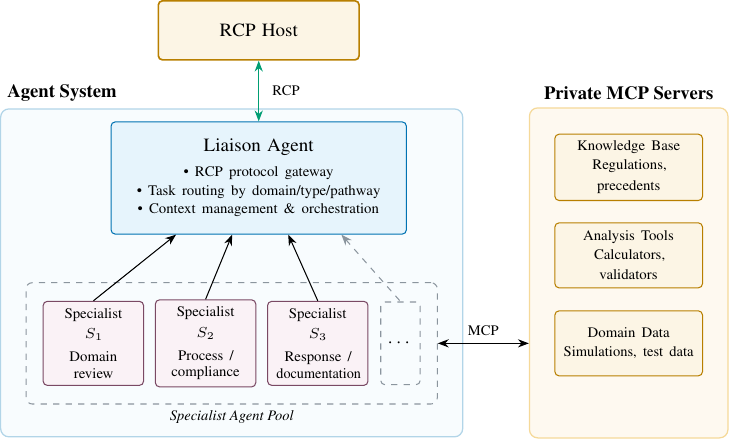}
    \caption{Agent architecture for the pilot implementation. Both agent systems adopt a common Liaison--Specialist topology: the Liaison Agent manages the RCP cross-boundary interface and routes tasks to domain-specific Specialists ($S_1 \ldots S_n$), which access private knowledge and analysis tools through an MCP Client.}
    \label{fig:agent_architecture}
\end{figure}

To exercise the RCP framework end-to-end, we implemented a pilot system mirroring an RAI cycle. RCP defines only the cross-boundary communication protocol; the internal organization of each party's agent system is left to the deploying organization. For this case study, both the regulator and the applicant were built around a common hierarchical Liaison--Specialist topology (\cref{fig:agent_architecture}). A Liaison Agent serves as the RCP protocol gateway and task orchestrator---managing external session state, authentication, and context tracking, and routing incoming requests to a pool of domain-specific Specialist Agents by technical domain, interaction type, and CFR pathway. The Liaison-to-Specialist hop is internal to the agent system and therefore outside the scope of RCP, which constrains only what crosses the trust boundary; the demonstration uses in-process orchestration via the Agent Development Kit (ADK), but an equally valid deployment could use A2A internally---without the RCP profile---or any other dispatch mechanism. Specialists access internal resources through an MCP Client that brokers calls to private knowledge bases and analysis tools. This separation of concerns lets specialist agents be tuned for their domain without carrying any of the external negotiation state, and it preserves a single-task abstraction at the trust boundary: the docket records one Liaison-owned Task per cross-boundary interaction, not the internal fan-out.

While the topology is shared, the two systems diverge in specialist composition. The NRC instantiation is review-oriented, with specialists for Neutronics, Thermal-Hydraulics, Structural, Environmental, Security, and QA/Licensing backed by MCP servers exposing CFR text, regulatory precedents, and analysis tools. The Applicant instantiation is response-oriented: a compliance validator maps internal design data to regulatory requirements, a document generator enforces submission formatting standards, and an RAI response specialist coordinates multi-domain answers. The asymmetry between reviewer-side and generator-side specialization---under an identical protocol surface---illustrates the kind of role flexibility the protocol is meant to admit.

\subsection{Demonstration}\label{sec:demonstration}

To validate the RCP framework, we executed a demonstration based on a hypothetical scenario regarding advanced reactor licensing. The scenario involves a mock NRC review of PRO-AID, a physics-informed digital twin for system-level fault detection and diagnosis~\cite{Nguyen2020Diagnostic, Nguyen2022DigitalTwin, dave2024integrating}, providing real-time condition monitoring on the Mechanisms Engineering Test Loop (METL) Sodium Purification System~\cite{Kultgen2016METL}. The application proposes three virtual sensor models---cold trap impurity capture rate, plugging temperature estimation, and electromagnetic (EM) pump efficiency---with Sequential Probability Ratio Test (SPRT) change detection, citing regulatory basis under 10~CFR~50.55a~\cite{CFR50_55a}, 10~CFR~50 Appendix~B~\cite{CFR50AppB}, 10~CFR~73.54~\cite{CFR73_54}, and RG~1.152~\cite{RG1_152}.

Both agent systems accessed domain knowledge through MCP servers: the NRC agent queried regulatory documents, while the Applicant agent accessed technical specifications and validation data. All exchanges were routed through the RCP Host, which enforced message typing, classification tagging, and immutable Context Stream logging. \Cref{fig:rai_workflow} presents the complete RAI lifecycle as a sequence diagram, illustrating how Liaison agents, the RCP Host, and private MCP servers coordinate across two phases: (1)~application submission and (2)~review and information exchange. The diagram then branches into two complementary outcomes: a \textit{lightweight information exchange} (Pattern~A) that resolves routine questions through repeatable request-response rounds, and a \textit{formal RAI cycle} (Pattern~B), triggered when lightweight exchange reveals a substantive compliance gap. In terms of the protocol's state machine (\cref{fig:task_fsm}), Pattern~A represents a loop between \texttt{working} and \texttt{input-required}, while Pattern~B requires transitioning through the \texttt{auth-required} gate. Per RCP semantics, each agent manages its own task state, while the RCP Host routes messages and persists them in the Context Stream without managing task state itself.

\subsubsection{Pattern~A: Lightweight Information Exchange}

\begin{figure}[!ht]
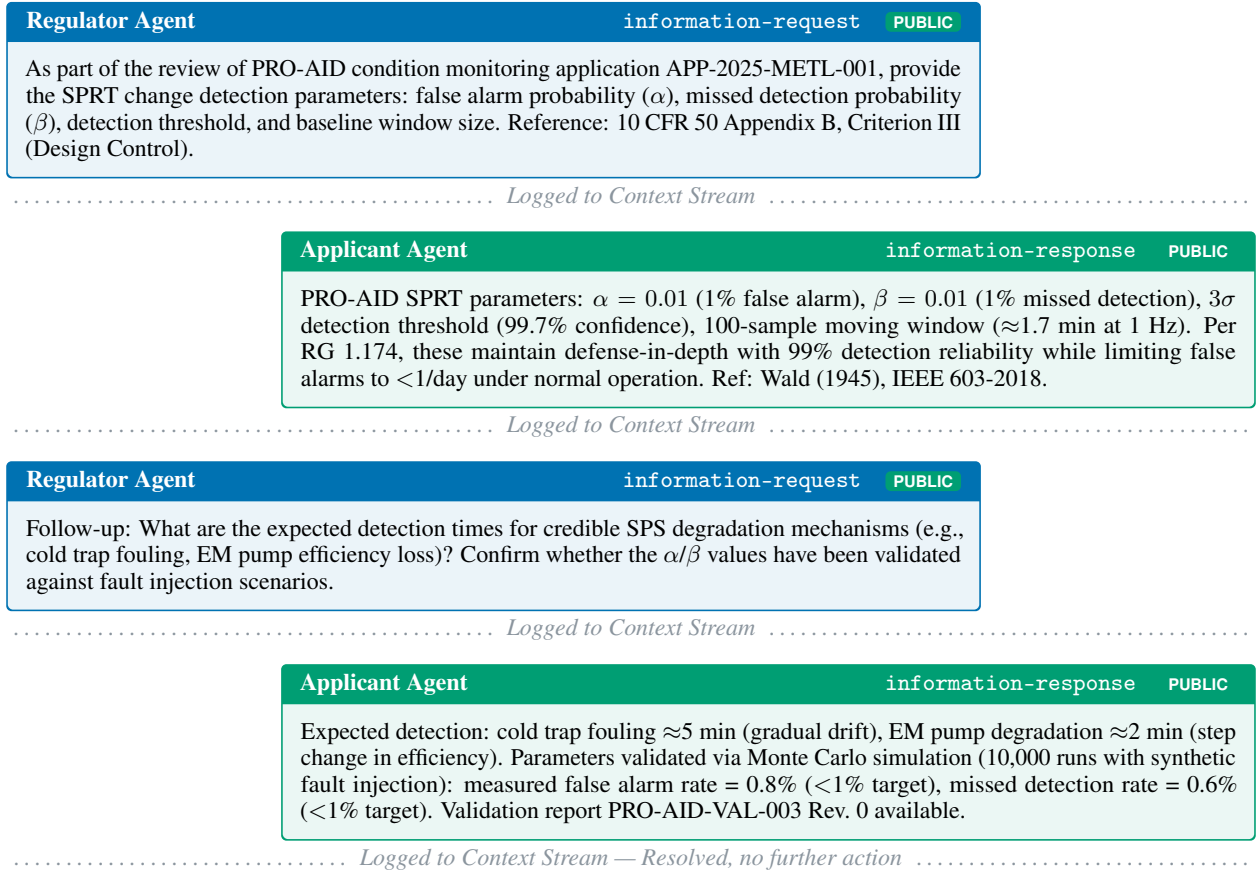


\regmsg{information-request}{\cpub}{%
As part of the review of PRO-AID condition monitoring application APP-2025-METL-001, provide the SPRT change detection parameters: false alarm probability ($\alpha$), missed detection probability ($\beta$), detection threshold, and baseline window size. Reference: 10~CFR~50 Appendix~B, Criterion~III (Design Control).}

\ledgerlog{Logged to Context Stream}

\appmsg{information-response}{\cpub}{%
PRO-AID SPRT parameters: $\alpha = 0.01$ (1\% false alarm), $\beta = 0.01$ (1\% missed detection), $3\sigma$ detection threshold (99.7\% confidence), 100-sample moving window ($\approx$1.7~min at 1~Hz). Per RG~1.174, these maintain defense-in-depth with 99\% detection reliability while limiting false alarms to $<$1/day under normal operation. Ref: Wald~(1945), IEEE~603-2018.}

\ledgerlog{Logged to Context Stream}

\regmsg{information-request}{\cpub}{%
Follow-up: What are the expected detection times for credible SPS degradation mechanisms (e.g., cold trap fouling, EM~pump efficiency loss)? Confirm whether the $\alpha$/$\beta$ values have been validated against fault injection scenarios.}

\ledgerlog{Logged to Context Stream}

\appmsg{information-response}{\cpub}{%
Expected detection: cold trap fouling $\approx$5~min (gradual drift), EM~pump degradation $\approx$2~min (step change in efficiency). Parameters validated via Monte Carlo simulation (10{,}000~runs with synthetic fault injection): measured false alarm rate~=~0.8\% ($<$1\% target), missed detection rate~=~0.6\% ($<$1\% target). Validation report PRO-AID-VAL-003~Rev.~0 available.}

\ledgerlog{Logged to Context Stream --- Resolved, no further action}

\caption{Pattern~A (lightweight information exchange): two rounds of bounded technical questions resolved entirely through \texttt{information-request}/\texttt{information-response} message pairs, all carrying the \protect\cpub{} classification. The NRC agent's follow-up is a natural consequence of the first response; both rounds complete without escalation to a formal RAI.}
\label{fig:patterna-resolved}
\end{figure}

\Cref{fig:patterna-resolved} presents a complete Pattern~A interaction spanning two request-response rounds. The NRC agent first requests specific SPRT implementation parameters to assess the adequacy of the PRO-AID change detection algorithm. Because the question is bounded---requesting factual design parameters rather than compliance justification---and the underlying methodology is published~\cite{Wald1945}, the Applicant agent retrieves the parameters from its knowledge base and responds directly, including a preemptive regulatory justification referencing RG~1.174 defense-in-depth principles. The NRC agent then issues a natural follow-up, asking for detection performance against credible degradation mechanisms. The Applicant's second response provides quantitative detection times and Monte Carlo validation results, fully satisfying the inquiry. Both rounds resolve without formal regulatory action.

In current NRC practice, even a straightforward parameter request like this would typically be consolidated into a formal RAI batch, triggering a multi-month response cycle involving management review. Under RCP, the \texttt{information-request}/\texttt{information-response} message pair provides a structured alternative for resolving bounded queries---including multi-turn follow-ups---while preserving full traceability through the Context Stream. As the ADAMS corpus analysis revealed (\cref{fig:adams_sankey}), 95\% of regulatory documents should proceed without blocking issues---suggesting that a substantial fraction of current RAI traffic consists of clarification questions amenable to Pattern~A resolution.

\subsubsection{Pattern~B: Escalation to Formal RAI}

\begin{figure}[!ht]
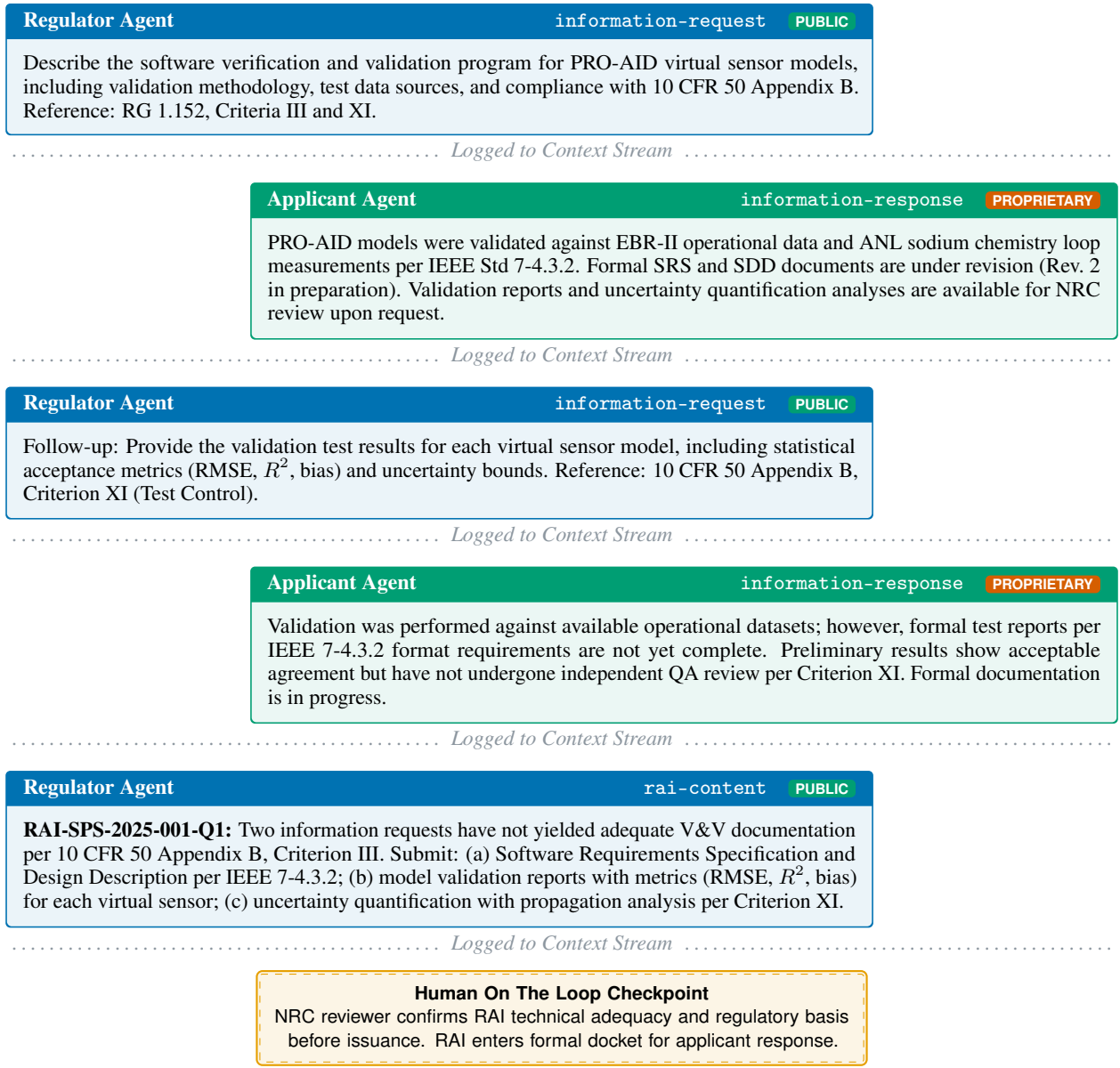


\regmsg{information-request}{\cpub}{%
Describe the software verification and validation program for PRO-AID virtual sensor models, including validation methodology, test data sources, and compliance with 10~CFR~50 Appendix~B. Reference: RG~1.152, Criteria~III and~XI.}

\ledgerlog{Logged to Context Stream}

\appmsg{information-response}{\cprop}{%
PRO-AID models were validated against EBR-II operational data and ANL sodium chemistry loop measurements per IEEE~Std~7-4.3.2. Formal SRS and SDD documents are under revision (Rev.~2 in preparation). Validation reports and uncertainty quantification analyses are available for NRC review upon request.}

\ledgerlog{Logged to Context Stream}

\regmsg{information-request}{\cpub}{%
Follow-up: Provide the validation test results for each virtual sensor model, including statistical acceptance metrics (RMSE, $R^2$, bias) and uncertainty bounds. Reference: 10~CFR~50 Appendix~B, Criterion~XI (Test Control).}

\ledgerlog{Logged to Context Stream}

\appmsg{information-response}{\cprop}{%
Validation was performed against available operational datasets; however, formal test reports per IEEE~7-4.3.2 format requirements are not yet complete. Preliminary results show acceptable agreement but have not undergone independent QA review per Criterion~XI. Formal documentation is in progress.}

\ledgerlog{Logged to Context Stream}

\regmsg{rai-content}{\cpub}{%
\textbf{RAI-SPS-2025-001-Q1:} Two information requests have not yielded adequate V\&V documentation per 10~CFR~50 Appendix~B, Criterion~III. Submit: (a)~Software Requirements Specification and Design Description per IEEE~7-4.3.2; (b)~model validation reports with metrics (RMSE, $R^2$, bias) for each virtual sensor; (c)~uncertainty quantification with propagation analysis per Criterion~XI.}

\ledgerlog{Logged to Context Stream}

\hitlcheck{NRC reviewer confirms RAI technical adequacy and regulatory basis before issuance. RAI enters formal docket for applicant response.}

\caption{Pattern~A to Pattern~B escalation: two rounds of lightweight exchange (Steps~1--4) fail to resolve a V\&V documentation gap, prompting the NRC agent to escalate to a formal RAI (Step~5). The HOTL checkpoint (\drref{4}) enforces human review before issuance. The figure terminates at RAI issuance: the formal response requires human-driven preparation and is intentionally outside the Applicant agent's scope. All exchanges are logged to the Context Stream.}
\label{fig:patternb-escalation}
\end{figure}

\begin{figure}[!htbp]
    \centering
    \hspace*{-1.25pt}\includegraphics{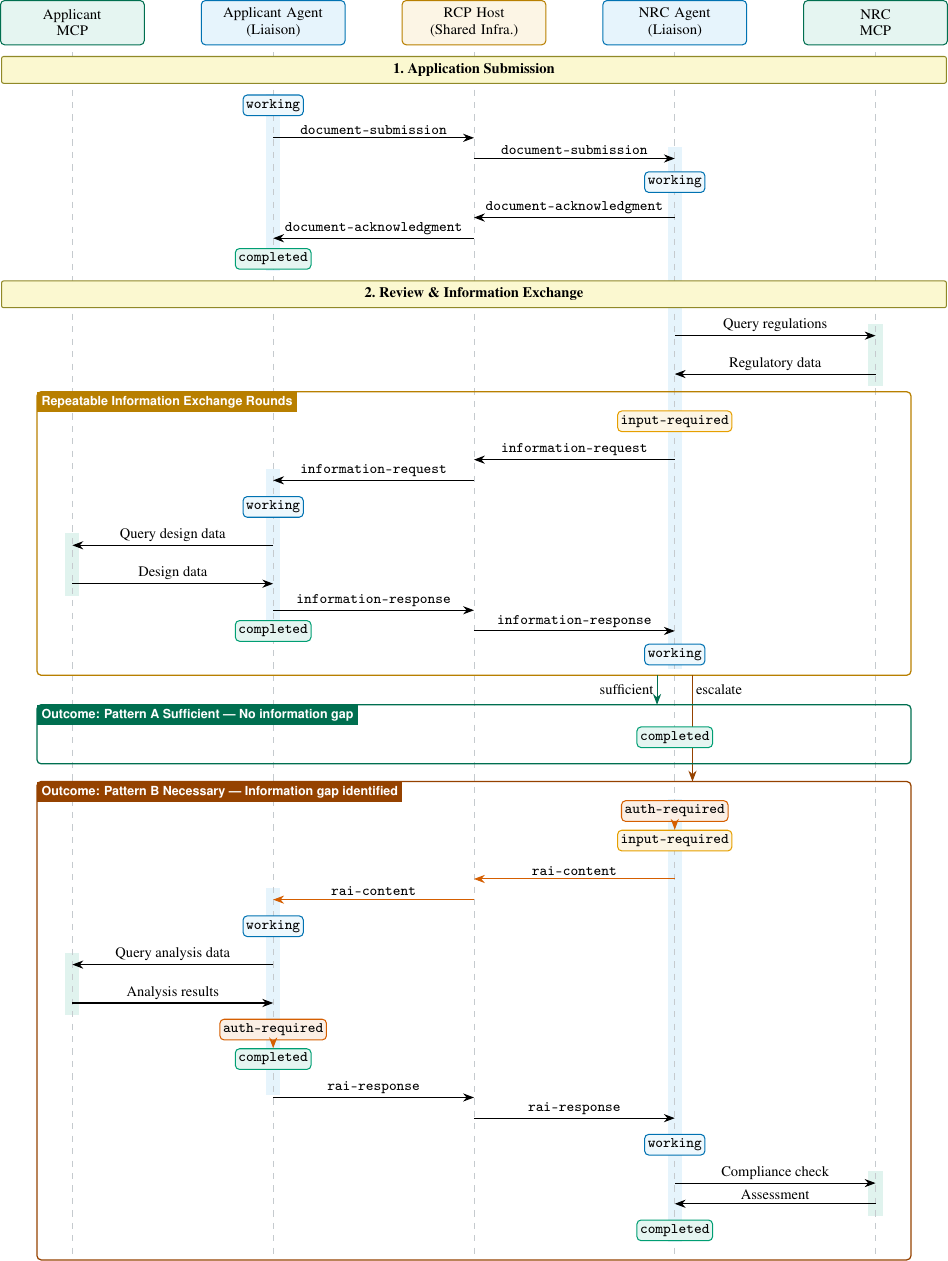}
    \caption{RCP RAI lifecycle showing the branching between Pattern~A (lightweight information exchange, \cref{fig:patterna-resolved}) and Pattern~B (formal RAI escalation, \cref{fig:patternb-escalation}). Each agent manages its own task state (\cref{fig:task_fsm}); state badges appear on the owning agent's lifeline. Arrow labels follow the message type catalog (\cref{tab:rcp_messages}). In Pattern~B the NRC agent enters \texttt{auth-required} for HOTL approval before issuing \texttt{rai-content}; exception paths are omitted for clarity.}
    \label{fig:rai_workflow}
\end{figure}

Not all questions can be resolved through lightweight exchange. \Cref{fig:patternb-escalation} presents a sequence in which the NRC agent attempts to resolve a question through two rounds of Pattern~A, determines that the Applicant's responses reveal a substantive compliance gap, and escalates to a formal RAI (Pattern~B). The exchange begins identically to \cref{fig:patterna-resolved}---an \texttt{information-request} about the software verification and validation program. The Applicant's initial response indicates that formal documentation is under revision, so the NRC agent issues a targeted follow-up requesting specific validation metrics. When this second response confirms that formal test reports have not undergone independent QA review, the agent escalates---having exhausted the lightweight channel. The figure deliberately terminates at the HOTL checkpoint and RAI issuance: the formal response requires human-driven engineering work, which is exactly the outcome the protocol is designed to produce when automated exchange reaches its limits.

Several RCP design features are visible in this exchange. First, the \textit{graduated escalation}: the NRC agent attempted two rounds of Pattern~A before determining that the compliance gap could not be resolved through lightweight exchange alone. The RAI text itself references this history (``Two information requests have not yielded adequate V\&V documentation''), providing a documented rationale for escalation. This graduated approach---try lightweight first, escalate only when necessary---is enabled by the protocol's typed separation of Pattern~A from Pattern~B and reinforced by the HOTL gate on \texttt{rai-content}, which makes escalation a deliberate, costly act. The protocol does not mandate the sequence, but its structure makes the cheap path the natural default. Second, the \textit{HOTL checkpoint} (\drref{4}) ensures that no formal RAI is issued without human specialist review, preserving regulatory authority while leveraging agentic efficiency for the investigative and drafting phases. Third, the \textit{data classification system} (\cref{sec:data_classification}) operates transparently throughout: the Applicant's \texttt{information-response} messages carry \cprop{} badges because they reference proprietary vendor-developed validation data and methodologies, while the NRC's messages here are \cpub{}, the default for routine regulatory correspondence on non-sensitive subject matter. The protocol enforces this classification structurally, routing proprietary content to the non-public portion of the docket.

While direct comparison between a demonstration and a production licensing review requires appropriate caution, the structural efficiency gains from protocol-driven exchange are clear: automated classification tagging replaces manual sensitivity review, immutable Context Stream logging replaces manual docketing, and---critically---Pattern~A deflection prevents routine questions from entering the formal RAI pipeline entirely. The contrast between \cref{fig:patterna-resolved} and \cref{fig:patternb-escalation} illustrates the core thesis: many regulatory interactions can be resolved through lightweight exchange, but when they cannot, the protocol provides a principled path to formal action---with a documented trail of prior attempts---that correctly hands off to human-driven processes.

\section{Results and Policy Implications}\label{sec:results_policy}

The demonstration presented in \cref{sec:demonstration} provides a basis for evaluating RCP against three dimensions relevant to regulatory reform: operational efficiency, transparency for oversight, and the handling of sensitive information. While a single demonstration does not constitute a comprehensive evaluation, the structural properties of the protocol enable grounded analysis of how each dimension would scale to production use.

\subsection{Efficiency Gains}

The primary efficiency contribution of RCP is the introduction of Pattern~A (lightweight information exchange) as a structured alternative to the formal RAI process. In current NRC practice, all substantive technical queries---whether simple parameter requests or complex compliance questions---follow the same formal pathway: staff draft, branch chief review, Office of the Executive Director for Operations (OEDO) routing, applicant response preparation, and formal docketing. This process typically requires 60--90~days per cycle~\cite{NuScaleLL2022}.

The METL PRO-AID demonstration illustrates how Pattern~A addresses bounded, factual queries before they enter this pipeline. The SPRT parameter exchange (\cref{fig:patterna-resolved}) resolved two rounds of technical inquiry---initial parameters and a follow-up on detection performance validation---in under two minutes total. Under current practice, these questions would be bundled into a formal RAI batch and processed over months. The efficiency gain is not merely temporal; it is structural. By providing a protocol-native channel for routine clarifications---including multi-turn follow-ups---RCP reduces the total volume of formal RAIs, allowing staff to concentrate review effort on substantive compliance gaps that genuinely require Pattern~B treatment.

The scale of this opportunity is suggested by the ADAMS corpus analysis (\cref{fig:adams_sankey}): 95\% of documents proceeded without blocking issues, and RAI responses dominated at 69\% of all interactions. Many of these addressed clarification questions---requests for specific design parameters, additional test data, or regulatory basis citations---that are structurally similar to the Pattern~A exchange demonstrated. If even a modest fraction of these could be resolved through lightweight exchange, the reduction in formal RAI volume would be significant. The NuScale DCA, which generated over 4,000~RAI questions and consumed more than \$500~million and 2~million labor hours~\cite{NuScaleCost}, represents the cost ceiling of fully-formal review processes. RCP's two-pattern architecture offers a path toward more proportional allocation of regulatory effort, where the formality of the process matches the complexity of the question.

To project these demonstration-level observations to a full design certification review, we develop a parametric cost and timeline model calibrated against the NuScale DCA scope~\cite{NuScaleLL2022} and the ADAMS corpus analysis presented in \cref{fig:adams_sankey}. The model compares three paradigms across cost and timeline dimensions:
\begin{itemize}
    \item Reconstructed Baseline \textbf{(RB)} --- conventional review, fully manual; both drafting and inter-organizational exchange are human-driven.
    \item Standalone Agents \textbf{(SA)} --- AI-assisted drafting within each organization, but the formal human-to-human inter-organizational pipeline is retained.
    \item \textbf{RCP} --- agent-to-agent communication over the protocol replaces the formal pipeline for routine exchanges; HOTL checkpoints remain on binding actions.
\end{itemize}

\subsubsection{Cost Model}

To quantify the financial impact of automating the regulatory review process, we must calculate the total review cost under the three distinct operational paradigms. We calculate these costs by estimating the total human labor hours required to resolve a projected volume of RAIs, scaling these hours by a blended labor rate, and applying paradigm-specific automation discounts.

To properly quantify effort required, each RAI is assigned to one of four \textit{complexity tiers} (indexed by $i$):
\begin{itemize}
    \item Routine --- factual queries: data fetch, status confirmation, retrieval of declared parameters.
    \item Standard --- structured analysis on established methods: section revisions, reconciliations, multi-turn clarification.
    \item Complex --- deep engineering analysis with multi-branch coordination (e.g., thermal-hydraulic transient reanalysis, structural reassessment).
    \item Novel --- first-of-a-kind methodologies and acceptance-criteria development; mandatory human specialist review.
\end{itemize}

The calculation is grounded in several empirically derived assumptions defining the fundamental review inputs. First, we assume a total volume of $N = 4{,}000$ RAIs, validated by historical data from the NuScale and ESBWR design certifications~\cite{NuScaleLL2022}. Second, we apply a blended labor rate of $r = \$250$/hr. This rate derives from the NRC full-cost hourly rate of \$318/hr (FY2025, 10~CFR~170.20)~\cite{FeeRule2025} and an estimated applicant loaded rate of approximately \$200/hr, yielding a 50/50 blend of approximately \$259/hr, which we round to \$250/hr. This is independently cross-validated by the NuScale DCA actuals: approximately \$500~million across 2~million labor hours~\cite{NuScaleLL2022}. Third, for each complexity tier $i$, the baseline human labor hours required to resolve a single RAI ($h_i$) and the inter-organizational process overhead fractions ($p_i$) are calibrated against NRC Cost Estimator data from approximately 5{,}000 licensing actions~\cite{NRCCostEstimator} and NRC workforce benchmarks~\cite{GAO071129}. The specific values for $h_i$, $p_i$, the computational API cost required to automate an RAI ($c_{\text{agentic},i}$), and the proportional share of RAIs falling into each tier ($f_i$) are summarized in \cref{tab:model_params}.

We first establish the RB cost ($C_{\text{RB}}$). This requires finding the weighted-average labor hours per RAI ($\bar{h}$) by summing the product of the tier proportions $f_i$ and the tier labor hours $h_i$. The RB cost is then the product of the total RAI volume $N$, the weighted-average labor hours $\bar{h}$, and the blended labor rate $r$:
\begin{align}
    \bar{h} &= \textstyle\sum_i f_i \, h_i \approx 89 \text{ hrs/RAI} \label{eq:weighted_hrs} \\
    C_{\text{RB}} &= \bar{h} r N \approx \$89\text{M} \label{eq:baseline}
\end{align}

To calculate the costs under the automated paradigms, we must introduce capability assumptions. We define the capability-dependent fraction of human labor displaced by AI automation under the RCP ($\alpha^{\text{RCP}}_i$) and SA ($\alpha^{\text{SA}}_i$) paradigms, which vary based on conservative, moderate, and optimistic scenarios fully detailed in \cref{tab:scenario_assumptions}. These values are author estimates calibrated against the published productivity-gain literature discussed in the surrounding text.

\begin{table}[htbp]
    \centering
    \caption{Per-tier automation rate assumptions across the three scenarios: conservative (Cons.), moderate (Mod.), and optimistic (Opt.).}
    \label{tab:scenario_assumptions}
    \small
    \begin{tabular}{l r r r r r r}
        \toprule
                 & \multicolumn{3}{c}{RCP, $\alpha^{\text{RCP}}_i$} & \multicolumn{3}{c}{SA, $\alpha^{\text{SA}}_i$} \\
        \cmidrule(lr){2-4} \cmidrule(lr){5-7}
        Tier     & Cons. & Mod. & Opt. & Cons. & Mod. & Opt. \\
        \midrule
        Routine  & 0.70 & 0.95 & 0.98 & 0.60 & 0.80 & 0.90 \\
        Standard & 0.63 & 0.85 & 0.92 & 0.40 & 0.60 & 0.75 \\
        Complex  & 0.53 & 0.72 & 0.80 & 0.25 & 0.40 & 0.55 \\
        Novel    & 0.40 & 0.55 & 0.65 & 0.15 & 0.25 & 0.35 \\
        \bottomrule
    \end{tabular}
\end{table}

Next, we calculate the cost under the RCP paradigm ($C_{\text{RCP}}$). For each tier, the absolute number of RAIs is determined by $N_i = f_i N$, yielding the total baseline human labor hours $H_{\text{human},i} = N_i h_i$. Because RCP structurally automates both drafting and inter-organizational routing, it displaces an $\alpha^{\text{RCP}}_i$ fraction of the total human labor across all tasks, leaving only the residual hours for human-in-the-loop review ($H_{\text{residual},i}$). The total cost rolls up across tiers by multiplying these residual hours by the labor rate $r$, and adding the computational API cost required to automate the tier's RAIs ($N_i c_{\text{agentic},i}$):
\begin{align}
    H_{\text{residual},i} &= H_{\text{human},i} (1 - \alpha^{\text{RCP}}_i) \label{eq:residual} \\
    C_{\text{RCP}} &= \textstyle\sum_i \left[ H_{\text{residual},i} r + N_i c_{\text{agentic},i} \right] \label{eq:rcp_cost}
\end{align}
Compute costs $c_{\text{agentic},i}$---based on current LLM API pricing~\cite{AnthropicPricing}---are negligible relative to labor at all tiers: even at the Novel tier, $c_{\text{agentic},i} = \$400$ represents less than 1.5\% of the \$27{,}000 residual labor cost.

Finally, we calculate the cost under the SA paradigm ($C_{\text{SA}}$). The structural difference between SA and RCP lies in the inter-organizational process overhead, represented by $p_i$. In the current paradigm, even when each organization deploys its own AI agent internally, communication must traverse a high-latency human-to-human channel (the red path in \cref{fig:rcp_flow}). SA automates internal analysis but cannot automate this inter-organizational pipeline. Therefore, the process-overhead share ($p_i H_{\text{human},i}$) is strictly excluded from the automatable base. Only the remaining internal fraction of the work, $(1 - p_i)$, is accessible to automation. This internal base is reduced by the internal automation rate ($\alpha^{\text{SA}}_i$), which may differ from the holistic RCP rate ($\alpha^{\text{RCP}}_i$) because a siloed agent faces a narrower displaceable task mix without structured data exchange. The residual hours for SA ($H_{\text{residual,SA},i}$) and total cost are calculated as:
\begin{align}
    H_{\text{residual,SA},i} &= H_{\text{human},i} \left[ 1 - (1 - p_i) \alpha^{\text{SA}}_i \right] \label{eq:standalone_auto} \\
    C_{\text{SA}} &= \textstyle\sum_i \left[ H_{\text{residual,SA},i} r + N_i c_{\text{agentic},i} \right] \label{eq:standalone_cost}
\end{align}

The central modeling claim is that RCP, by replacing the human-to-human channel with the low-latency agent-to-agent path (the green path in the bottom panel of \cref{fig:rcp_flow}), makes the process-overhead hours themselves automatable. The full labor base thus enters \cref{eq:residual} for RCP, whereas only the non-overhead portion enters \cref{eq:standalone_auto} for SA, forming the basis of the cost gap between the two paradigms.

\begin{table}[htbp]
    \centering
    \caption{Tier-level model inputs derived from the ADAMS corpus analysis (\cref{sec:corpus_analysis}).}
    \label{tab:model_params}
    \small
    \begin{tabular}{l r r r r}
        \toprule
             & Corpus       & Hrs/       & Compute                      & Process         \\
        Tier & Share, $f_i$ & RAI, $h_i$ & /RAI, $c_{\text{agentic},i}$ & Overhead, $p_i$ \\
        \midrule
        Routine  & 12\% &   2 &   \$5 & 50\% \\
        Standard & 24\% &  12 &  \$50 & 35\% \\
        Complex  & 57\% & 120 & \$200 & 25\% \\
        Novel    &  7\% & 240 & \$400 & 20\% \\
        \bottomrule
    \end{tabular}
\end{table}

\subsubsection{Timeline Model}

To determine the schedule compression achieved by automating the regulatory process, we must calculate the overall review timeline under the RCP and SA paradigms. We calculate this duration by summing the time required to process parallel batches of independent RAIs and the time consumed by serial chains of dependent RAIs, identifying the tier that governs the critical path.

The timeline calculation relies on modeling the shared internal processing times alongside the paradigm-specific inter-organizational overheads. First, we assume that both SA and RCP use identical agentic drafting and HOTL review pipelines. For any single cycle, the agentic drafting time ($T_{\text{draft}}$) runs concurrently with any necessary physics computation ($T_{\text{comp}}$---such as RELAP5 or CFD analysis, which cannot be compressed by LLMs). Thus, the effective processing time ($T_{\text{eff}}$) forms the internal bottleneck: $T_{\text{eff}} = \max(T_{\text{draft}}, T_{\text{comp}})$. The total time required to clear the HOTL review queue for a parallel batch ($T_{\text{HOTL}}$) is bound by the minimum cognitive review time per draft ($T_{\text{base}}$) and the throughput of the Subject Matter Expert (SME) review queue in reviews per day ($B$), formalized as $T_{\text{HOTL}} = \max(T_{\text{base}},\, N_{\text{parallel},i} / B)$. 

Second, the timeline depends heavily on how RAIs are sequenced and routed between organizations. The unblocked fractions determining the number of parallelizable ($N_{\text{parallel},i}$) and blocked or serial ($N_{\text{serial},i}$) RAIs per complexity tier, as well as the depth of the serial dependency chains ($D$), are empirically derived from our ADAMS corpus analysis. Third, we define the inter-organizational processing overhead required to route these batches. For the SA paradigm, the formal processing overhead ($T_{\text{proc}}^{\text{SA}}$) varies by tier and is significantly longer due to manual human handoffs. For the RCP paradigm, the automated agent-to-agent routing overhead ($T_{\text{proc}}^{\text{RCP}}$) is a constant 5~days across all tiers. \cref{tab:time_params} summarizes the specific empirical values assigned to these variables for each tier.

With the structural parameters defined, we calculate the time required for parallel batches ($T_{\text{parallel}}^P$) and serial chains ($T_{\text{serial}}^P$). Under both paradigms, the mathematical structure is identical. Letting $P \in \{\text{RCP}, \text{SA}\}$ denote the operational paradigm, the duration of the parallel batch is the sum of the effective processing time, the total HOTL review time, and the paradigm's specific routing overhead. The serial duration is constrained by the dependency chain depth ($D$), scaling multiplicatively with the duration of a single serial cycle (effective processing, base human review, and routing overhead):
\begin{align}
    T_{\text{parallel}}^P &= T_{\text{eff}} + T_{\text{HOTL}} + T_{\text{proc}}^P \label{eq:t_par} \\
    T_{\text{serial}}^P   &= D (T_{\text{eff}} + T_{\text{base}} + T_{\text{proc}}^P) \label{eq:t_ser}
\end{align}

The total timeline ($T_{\text{total}}^P$) is then calculated by identifying the single complexity tier that forms the critical path (the maximum sum of parallel and serial times across all tiers $i$), applying a conversion factor $k = 30$~days/month, and adding a fixed pre-application overhead $T_{\text{pre}} = 6$~months:
\begin{align}
    T_{\text{total}}^P &= \frac{1}{k} \max_i\!\left(T_{\text{parallel},i}^P + T_{\text{serial},i}^P\right) + T_{\text{pre}} \label{eq:t_total}
\end{align}

\begin{table}[htbp]
    \centering
    \caption{Time model parameters by tier. All symbols are defined in the surrounding text. SME throughput $B$ is estimated from NRC organizational capacity (``NRC at a Glance'')~\cite{NUREG1350}; the unblocked fraction is from the ADAMS corpus analysis (\cref{sec:corpus_analysis}).}
    \label{tab:time_params}
    \small
    \begin{tabular}{l r r r r r r r r}
        \toprule
        Tier & $T_{\text{draft}}$ & $T_{\text{comp}}$ & $T_{\text{base}}$ & $T_{\text{proc}}^{\text{SA}}$ & $T_{\text{proc}}^{\text{RCP}}$ & $B$ (rev/d) & $D$ & Unblocked \\
        \midrule
        Routine  &  5 min & ---    & ---  & 15 d & 5 d & 100  & 1 & 100\% \\
        Standard & 10 min & 0.5 d  & 1 d  & 20 d & 5 d &  30  & 2 &  98\% \\
        Complex  &  1 hr  &  5 d   & 3 d  & 25 d & 5 d & 12.5 & 5 &  93\% \\
        Novel    &  6 hr  & 15 d   & 7 d  & 30 d & 5 d &  3.0 & 6 &  87\% \\
        \bottomrule
    \end{tabular}
\end{table}

The Novel tier dominates this critical path calculation. With $T_{\text{comp}} = 15$~d, $T_{\text{base}} = 7$~d, $D = 6$, and 87\% unblocked, the 291 Novel RAIs split into 252 parallel and 39 serial items. The effective time is 15~days, and the HOTL queue clears in 84~days at a throughput of 3.0~reviews/day. Evaluating the equations yields $T_{\text{total}}^{\text{RCP}} = (104 + 162)/k + T_{\text{pre}} \approx 15$~months versus $T_{\text{total}}^{\text{SA}} = (129 + 312)/k + T_{\text{pre}} \approx 21$~months, resulting in a gap of approximately 6~months. Structurally, this gap scales as $D (T_{\text{proc}}^{\text{SA}} - T_{\text{proc}}^{\text{RCP}})$, explaining why timeline compression is highly dependent on replacing the inter-organizational communication channel rather than the underlying automation rate.

\subsubsection{Projected Savings}

To translate the preceding mathematical models into projected macro-level savings, we evaluate the cost (\cref{eq:rcp_cost,eq:standalone_cost}) and timeline (\cref{eq:t_total}) equations across the full conservative-to-optimistic capability envelope. By systematically applying the projected $\alpha^{\text{RCP}}_i$ and $\alpha^{\text{SA}}_i$ automation rates alongside paradigm-specific routing overheads ($T_{\text{proc}}^P$) to the empirical baseline derived from the ADAMS corpus, we can isolate the structural advantages of protocol-driven communication. \cref{tab:paradigm_comparison} summarizes these aggregated cost and schedule impacts for a full 4{,}000-RAI design certification review.

\begin{table}[htbp]
    \centering
    \caption{Cost and timeline comparison across three automation scenarios. The baseline is the RB. Timeline compression is driven by inter-organizational routing efficiencies rather than drafting automation rates, remaining constant across scenarios.}
    \label{tab:paradigm_comparison}
    \small
    \begin{tabular}{l l r r r r}
        \toprule
        Paradigm & Scenario & Cost & Savings & Timeline & Reduction \\
        \midrule
        RB & Baseline & ${\sim}$\$89M & ---  & 42 mo        & ---  \\
        \midrule
        \multirow{3}{*}{SA} 
                 & Cons.    & ${\sim}$\$73M & 17\% & \multirow{3}{*}{${\sim}$21 mo} & \multirow{3}{*}{50\%} \\
                 & Mod.     & ${\sim}$\$64M & 28\% & & \\
                 & Opt.     & ${\sim}$\$54M & 38\% & & \\
        \midrule
        \multirow{3}{*}{RCP} 
                 & Cons.    & ${\sim}$\$44M & 50\% & \multirow{3}{*}{${\sim}$15 mo} & \multirow{3}{*}{65\%} \\
                 & Mod.     & ${\sim}$\$28M & 68\% & & \\
                 & Opt.     & ${\sim}$\$21M & 77\% & & \\
        \bottomrule
    \end{tabular}
\end{table}

\begin{figure}[htbp]
    \centering
    \hspace*{-1.25pt}\includegraphics{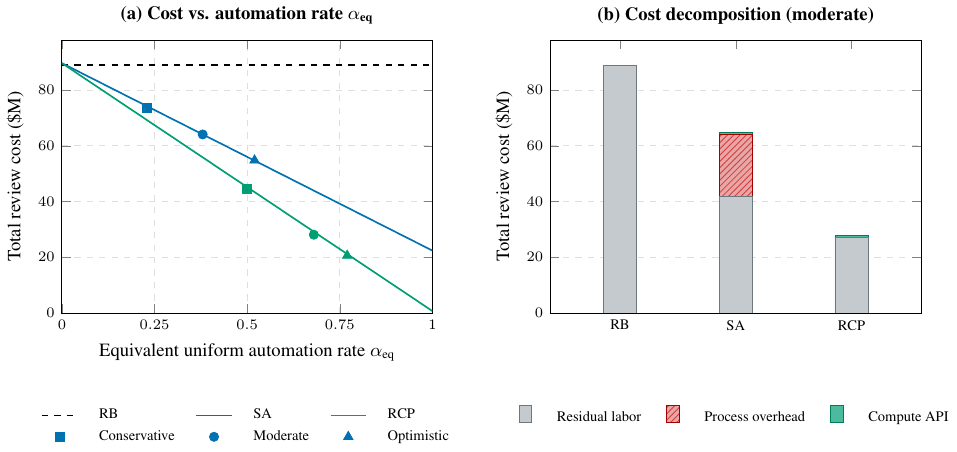}
    \caption{Cost dominance of RCP over SA, viewed two ways. \textbf{(a)}~Total review cost as a function of a single equivalent uniform automation rate $\alpha_{\text{eq}}$ applied to every tier---a one-parameter collapse of the per-tier $\alpha^P_i$ for visualization. RCP automates the full labor base while SA can only automate the $(1-p_i)$ non-overhead share, so the RCP curve sits strictly below SA at every $\alpha_{\text{eq}}$; the gap is structural, not a capability claim. Markers locate the conservative, moderate, and optimistic scenarios (\cref{tab:scenario_assumptions}) at the equivalent uniform $\alpha_{\text{eq}}$ that reproduces each scenario's per-tier cost. \textbf{(b)}~Decomposition of the moderate-scenario cost. The red hatched block is the inter-organizational process overhead ($p_i H_{\text{human},i}$); SA must pay it as a structural floor, while RCP eliminates it by replacing the formal pipeline with agent-to-agent routing.}
    \label{fig:pareto_frontier}
\end{figure}

Under the moderate scenario, RCP saves approximately \$36M more than SA and compresses the timeline by 6~additional months---a total reduction of 27~months (65\%) and 68\% in cost relative to the RB. Notably, the table reveals that RCP's \textit{conservative} scenario (\$44M) significantly outperforms SA's \textit{optimistic} scenario (\$54M). The cost gap arises because RCP can automate the process overhead that SA must preserve; the time gap arises because RCP eliminates per-cycle formal pipeline delays that compound across serial dependency chains. \cref{fig:pareto_frontier} visualizes both perspectives on this gap. Panel~(a) sweeps a single equivalent uniform automation rate $\alpha_{\text{eq}}$\footnote{The equivalent uniform rate $\alpha_{\text{eq}}$ for each scenario is calculated as the baseline-labor-hour-weighted average of its per-tier automation assumptions: $\alpha_{\text{eq}} = \sum_i (H_{\text{human},i} \, \alpha^P_i) / \sum_i H_{\text{human},i}$, where $P \in \{\text{RCP}, \text{SA}\}$ denotes the operational paradigm.} to demonstrate that RCP's structural cost advantage holds continuously across the full capability spectrum. Panel~(b) then decomposes the moderate-scenario cost to explicitly expose the inter-organizational process-overhead block that acts as an un-automatable floor for SA. \cref{tab:cost_breakdown} subsequently disaggregates these savings by complexity tier. The Complex tier dominates, contributing 80\% of total cost savings under the moderate scenario, due to its high corpus share (57\%) and substantial automation potential.

\begin{table}[htbp]
    \centering
    \caption{Cost savings by complexity tier under the moderate scenario. Compute costs are negligible ($<$\$1M total) and are included in the RCP Cost.}
    \label{tab:cost_breakdown}
    \small
    \begin{tabular}{l r r r r}
        \toprule
        Tier & RB Cost & RCP Cost & Savings & Hrs Saved \\
        \midrule
        Routine  &   \$0.2M &  \$0.0M &  \$0.2M &     885 \\
        Standard &   \$2.9M &  \$0.5M &  \$2.4M &   9{,}802 \\
        Complex  &  \$68.4M & \$19.6M & \$48.8M & 197{,}078 \\
        Novel    &  \$17.5M &  \$8.0M &  \$9.5M &  38{,}412 \\
        \midrule
        \textbf{Total} & \textbf{\$89M} & \textbf{\$28M} & \textbf{\$61M} & \textbf{246{,}178} \\
        \bottomrule
    \end{tabular}
\end{table}

The automation assumptions driving these cost projections are firmly grounded in empirical studies of AI-assisted knowledge work. The conservative and optimistic bounds defined in \cref{tab:scenario_assumptions} span the range of productivity gains observed across multiple industries. Under conservative assumptions, RCP achieves approximately \$44M in cost (50\% savings), a floor that aligns with controlled field experiments by Dell'Acqua et al.~\cite{DellAcqua2023}, who found AI users 25\% faster with 32\% higher quality on consulting tasks, and with BCG's reported 30--50\% acceleration of business processes for AI-enabled enterprises~\cite{BCG_AI2025}. The moderate estimate (68\% savings, \$28M) is consistent with McKinsey's projection that generative AI could automate 60--70\% of knowledge work activities~\cite{McKinseyAI2023}. Meanwhile, the GitHub Copilot study by Peng et al.~\cite{Peng2023}, which measured a 55.8\% speedup for developers on structured coding tasks---the closest analogue to structured RAI response drafting---falls cleanly between the conservative and moderate scenarios. For additional context, the NRC's own process improvements without AI have achieved 35\% reductions in staff hours and 40\% in schedules for operating reactor licensing actions~\cite{NRCEfficiency2025}, suggesting that significant further gains require the structural automation RCP provides.

Beyond financial impact, the projected schedule compression satisfies both industry targets and recent legislative mandates for advanced reactor reviews. The 15-month RCP timeline comfortably meets independent benchmarks: the Nuclear Energy Institute's recommended maximum of 18~months for first-of-a-kind advanced reactor licensing~\cite{NEI2025}, the ADVANCE Act's statutory mandate~\cite{ADVANCEAct2024}, and Executive Order~14300's requirement that the NRC complete certain reviews within 18~months~\cite{EO14300}. Because this timeline is bottlenecked by SME-queue throughput and per-cycle processing delays rather than by per-tier automation rates, the 15-month projection holds across the entire conservative-to-optimistic envelope. The model thus demonstrates that RCP can meet the 18-month target with margin under any capability scenario, despite not being explicitly calibrated to those external goals. Notably, the projected percentage savings are also robust against proportional scaling of the hours-per-RAI estimates: varying the weighted average by $\pm$25\% changes the absolute dollar figures but preserves the 50--77\% reduction ratios, as the key driver remains the automation rate rather than the baseline labor estimate. Ultimately, by decoupling review throughput from the linear constraint of human-driven inter-organizational routing, RCP offers a structural solution to the ``crisis of scale'' introduced in Section~1.

\subsection{Transparency and Oversight}

A central design requirement of RCP is that agentic automation must not diminish---and should actively enhance---the transparency of regulatory proceedings. The demonstration provides evidence that this requirement is achievable.

Every exchange in both demonstration scenarios was logged to the Context Stream---defined in \cref{sec:framework} as the protocol's continuous, append-only ledger. This ledger captures message content, classification, timestamps, and participant identifiers, producing a machine-readable docket that is structurally richer than current ADAMS entries, which are typically unstructured PDF documents uploaded after the fact. Under RCP, as illustrated by the logged steps in \cref{fig:patterna-resolved} and \cref{fig:patternb-escalation}, the Context Stream provides an indexed, queryable record of every interaction. This enables real-time oversight by authorized personnel and retrospective reconstruction of the complete proceeding history.

Beyond the external exchange record on the shared RCP Host, the NRC agent's internal reasoning process remains auditable within the agency's private zone through MCP activity logs. During the application review that led to the formal RAI (\cref{fig:patternb-escalation}), the agent executed five knowledge base queries against internal regulatory documents, searching for requirements related to ``digital twin condition monitoring,'' ``ASME Section~XI alternatives,'' ``software verification and validation,'' and ``Appendix~B quality assurance criteria.'' Each query returned ranked results with relevance scores, and the internal trace documents why software V\&V was identified as the most critical gap---specifically, that the application cited RG~1.152 compliance but provided no evidence of validation testing, a finding grounded in 10~CFR~50 Appendix~B Criterion~III. This internal reasoning chain---from knowledge base query to gap identification to draft RAI formulation---is preserved in its entirety, satisfying the Epistemic Grounding requirement (\drref{3}) by making the citation basis of every regulator-facing claim internally auditable.

Crucially, structural transparency does not imply immediate public disclosure. While the external Context Stream and internal reasoning traces create a comprehensive record, access to these logs remains under strict agency purview. Public release continues to be governed by the NRC's established classification rules, FOIA procedures, and sensitive information controls. However, when disclosure is legally required---such as during ASLB proceedings or ex parte review under 10~CFR~2.347 and~2.348---the protocol provides the exact, discoverable record demanded by \drref{2}. Furthermore, this internal transparency supports the High-Impact AI oversight requirements of OMB Memorandum~M-25-21~\cite{OMBM2521}. By capturing a reviewer's reasoning in real time rather than in a summary Safety Evaluation Report written months later, the protocol transforms the docket from a retrospective documentation exercise into a continuous, machine-verifiable record of regulatory reasoning.

\subsection{Privacy and Security}\label{sec:privacy_security}

The handling of sensitive information---proprietary data, SGI, and CUI---is a first-order concern in regulatory communication. The RCP architecture addresses this through two complementary mechanisms: classification tagging at the message level (\cref{sec:data_classification}) and privacy-by-design separation at the system level (\cref{sec:protocol_spec}).

In the demonstration, every message carried an explicit classification badge (\cpub{} or \cprop{}), enforced by the protocol rather than left to manual annotation. The Applicant's \texttt{information-response} messages in \cref{fig:patternb-escalation} illustrate how this operates in practice: these responses reference proprietary validation methodologies and vendor-developed algorithms, triggering \cprop{} classification. Asymmetric classification occurs naturally within the workflow: an NRC request may rely entirely on public regulatory criteria (\cpub{}), while the Applicant's response containing specific design details is restricted (\cprop{}). Conversely, if the NRC's inquiry references proprietary applicant data or pre-decisional internal analysis, the agent inherits the appropriate restrictive classification. The protocol requires that proprietary content be accompanied by withholding justification under 10~CFR~2.390(a)(4)---mirroring the existing regulatory requirement for affidavits of confidentiality---but enforces this requirement structurally rather than procedurally. A message marked \cprop{} is automatically routed to the non-public portion of the Context Stream; a message without a valid withholding basis cannot receive proprietary classification. This eliminates the class of errors in which sensitive information is inadvertently exposed in the public record---a risk that currently requires manual quality checks by document control staff.

At the architectural level, the privacy-by-design separation (\cref{fig:component_arch}) ensures that each organization's internal knowledge---accessed through MCP servers---never crosses the trust boundary. The NRC agent's regulatory knowledge base and the Applicant's proprietary design data remain within their respective private zones. Only the typed RCP messages---whose content and classification are explicitly chosen by the sending agent---traverse the shared RCP Host infrastructure. This structural guarantee is stronger than current practice, where inadvertent disclosure of pre-decisional or proprietary information through email attachments or misfiled ADAMS entries remains a documented risk.

\subsection{Generalizing RCP Across Regulatory Interaction Modes}\label{sec:audits}

The preceding analysis treats the RAI as the canonical formal exchange, but the same protocol mechanics apply to a second, increasingly prominent mode of regulator--applicant interaction: the regulatory audit. Whereas an RAI is a formal, docketed request for information, an audit is an interactive examination of primarily \textit{non-docketed} information---calculations, design-basis documents, software codes, and licensee-retained records---conducted to gain understanding and to identify what must ultimately be docketed to support a safety finding~\cite{NRCLIC111}. The NRC has moved toward this mode, encouraging staff to use regulatory audits to improve review efficiency~\cite{NRCLIC111}, and industry assessments report that audits both reduce the number of RAIs issued and compress review schedules through more frequent, direct communication~\cite{NIALicensing2023}. This is a complement to the RAI process rather than a replacement: any information the staff relies upon for its determination must still be formally docketed---through an RAI or voluntary submission---so the two channels coexist.

The same efficiency push driving regulatory audits has also produced lighter-weight formal instruments, most notably the Request for Confirmatory Information (RCI), reserved for low-complexity, high-confidence, factual matters resolved by a short confirmatory response~\cite{NRCStreamlining}. Together with the RAI, these instruments span a spectrum of formality that the protocol's two-pattern workflow (\cref{sec:demonstration}) would naturally accommodate. The RCI maps cleanly onto Pattern~A, and an audit---though a broader, more open-ended activity---is in practice conducted as a sequence of such lightweight, iterative exchanges rather than a single formal request; the heavyweight RAI, by contrast, follows the escalation pathway of Pattern~B. Because all of these interactions fundamentally rely on the secure exchange of information across an organizational trust boundary, RCP is a natural fit, transforming the NRC's emerging virtual audits~\cite{NRCStreamlining} from an ad hoc portal arrangement into a classification-aware, fully auditable channel.

\section{Concluding Remarks}\label{sec:conclusion}

We have presented RCP as a standardized framework to address the structural bottlenecks inherent in modern administrative review. The core design of this protocol was inspired by our empirical analysis of the nuclear regulatory corpus, which revealed that 95\% of docketed interactions proceed without external blocking issues---identifying a massive opportunity to replace the formal human-to-human pipeline with a structured, agent-to-agent communication channel that spans the full spectrum of regulatory interaction---from lightweight confirmatory exchanges and audits to formal RAIs. By evaluating this shift across conservative to optimistic capability scenarios and anchoring our assumptions in the published literature on AI productivity gains, the magnitude of the opportunity becomes clear. Against an RB of \$89M and 42~months, this structural shift projects a 65\% compression in review schedules (to 15~months) and upwards of a 77\% reduction in direct labor costs (down to \$21M).

While these efficiency gains are compelling, realizing this shift requires that both applicants and regulators entrust sensitive technical data to a shared protocol infrastructure. This step demands not only robust technical safeguards---such as the privacy-by-design architecture ensuring Information Sovereignty (\drref{1}) and the cryptographically Verifiable Shared State (\drref{2}) described in this work---but also significant institutional willingness to adopt fundamentally new modes of interaction. For agencies like the NRC, this means accepting machine-mediated communication, backed by Epistemic Grounding (\drref{3}) and continuous Human Oversight (\drref{4}), as part of the official, auditable docket record; for applicants, it means exposing proprietary design data to an agentic pipeline. 

Neither is a purely technical decision. Adoption will require regulatory rulemaking, updated staff guidance, and sustained leadership commitment from both industry and government---political will that historically lags the technology it governs. These institutional barriers, rather than any algorithmic or architectural limitation, are likely the primary obstacle to deployment. Yet, the cost of inaction is severe. The friction of inter-organizational exchange is not unique to nuclear licensing; the same bottleneck characterizes pharmaceutical approvals, environmental permitting, and financial supervision. With the US regulatory paperwork burden carrying an estimated \$426.5~billion annual opportunity cost, replicating the projected 50--77\% structural efficiencies across the administrative state implies savings on the order of \$210--\$330~billion per year---approaching 1\% of US GDP.

\section*{CRediT authorship contribution statement}
\textbf{Akshay J.~Dave}: Conceptualization, Methodology, Software, Visualization, Writing - Original Draft.
\textbf{David Grabaskas:} Funding acquisition, Investigation, Writing - Review \& Editing.
\textbf{Joseph A.~Renevitz:} Funding acquisition, Writing - Review \& Editing.
\textbf{Richard B.~Vilim:} Funding acquisition, Writing - Review \& Editing.

\section*{Acknowledgements}
This work was supported by the DOE Regulatory Framework Modernization program.

\bibliographystyle{unsrtnat}
\bibliography{references}

\end{document}